\documentclass{article}

\usepackage[final,nonatbib]{neurips_2021}

\usepackage[utf8]{inputenc} 
\usepackage[T1]{fontenc}    
\usepackage[bookmarks=true]{hyperref}       
\usepackage{url}            
\usepackage{booktabs}       
\usepackage{amsfonts}       
\usepackage{nicefrac}       
\usepackage{microtype}      
\usepackage{xcolor}         

\usepackage{times}

\usepackage[numbers]{natbib}
\usepackage{multicol}
\usepackage{xspace}
\usepackage{amsmath}
\usepackage{amssymb}

\usepackage{siunitx}
\usepackage{graphicx}
\usepackage{subcaption}
\usepackage{booktabs}       

\usepackage{mathtools}

\DeclarePairedDelimiter\abs{\lvert}{\rvert}%

\usepackage{comment}

\usepackage{algorithm} 
\usepackage{algorithmic}  
\usepackage[algo2e,linesnumbered]{algorithm2e} 
\usepackage{bm}

\usepackage{xcolor}

\let\labelindent\relax
\usepackage{enumitem}

\usepackage{algorithm2e}

\newcommand{\xxnote}[3]{}
\ifx\hidenotes\undefined
  \usepackage{color}
  \renewcommand{\xxnote}[3]{\color{#2}{#1: #3}}
\fi

\title{Differentiable Quality Diversity}

\author{
    Matthew C. Fontaine \\
    University of Southern California \\
    Los Angeles, CA \\
    mfontain@usc.edu
    
    \And
    
    Stefanos Nikolaidis \\ 
    University of Southern California \\
    Los Angeles, CA \\
    nikolaid@usc.edu
}

\begin{document}

\maketitle

\begin{abstract}

Quality diversity (QD) is a growing branch of stochastic optimization research that studies the problem of generating an archive of solutions that maximize a given objective function but are also diverse with respect to a set of specified measure functions. However, even when these functions are differentiable, QD algorithms treat them as \mbox{``black boxes’’}, ignoring gradient information. We present the differentiable quality diversity (DQD) problem, a special case of QD, where both the objective and measure functions are first order differentiable. We then present MAP-Elites via a Gradient Arborescence (MEGA), a DQD algorithm that leverages gradient information to efficiently explore the joint range of the objective and measure functions. Results in two QD benchmark domains and in searching the latent space of a StyleGAN show that MEGA significantly outperforms state-of-the-art QD algorithms, highlighting DQD's promise for efficient quality diversity optimization when gradient information is available. Source code is available at \url{https://github.com/icaros-usc/dqd}.

\end{abstract}

\section{Introduction}

We introduce the problem of differentiable quality diversity (DQD) and propose the \mbox{MAP-Elites} via a Gradient Arborescence (MEGA) algorithm as the first DQD algorithm.

Unlike single-objective optimization, quality diversity (QD) is the problem of finding a range of high quality solutions that are diverse with respect to prespecified metrics. For example, consider the problem of generating realistic images that match as closely as possible a target text prompt \mbox{``Elon Musk''}, but vary with respect to hair and eye color. We can formulate the problem of searching the latent space of a generative adversarial network (GAN)~\citep{goodfellow:nips14} as a QD problem of discovering latent codes that generate images maximizing  a matching score for the prompt ``Elon Musk'', while achieving a diverse range of measures of hair and eye color, assessed by visual classification models~\cite{radford2021learning}. More generally, the QD objective is to maximize an objective $f$  for each output combination of measure functions $m_i$. A QD algorithm produces an archive of solutions, where the algorithm attempts to discover a representative for each measure output combination, whose $f$ value is as large as possible.

While our example problem can be formulated as a QD problem, all current QD algorithms treat the objective $f$ and measure functions $m_i$ as a black box. This means, in our example problem, current QD algorithms fail to take advantage of the fact that both $f$ and $m_i$ are end-to-end differentiable neural networks. Our proposed differentiable quality diversity (DQD) algorithms leverage first-order derivative information to significantly improve the computational efficiency of solving a variety of QD problems where $f$ and $m_i$ are differentiable.

To solve DQD, we introduce the concept of a \textit{gradient arborescence}. Like gradient ascent, a gradient arborescence makes greedy ascending steps based on the objective $f$. Unlike gradient ascent, a gradient arborescence encourages exploration by branching via the measures $m_i$. We adopt the term \textit{arborescence} from the minimum arborescence problem~\citep{chu:ss65} in graph theory, a directed counterpart to the minimum spanning tree problem, to emphasize the directedness of the branching search.

\begin{figure}[!t]
\centering
\includegraphics[width=0.9\linewidth]{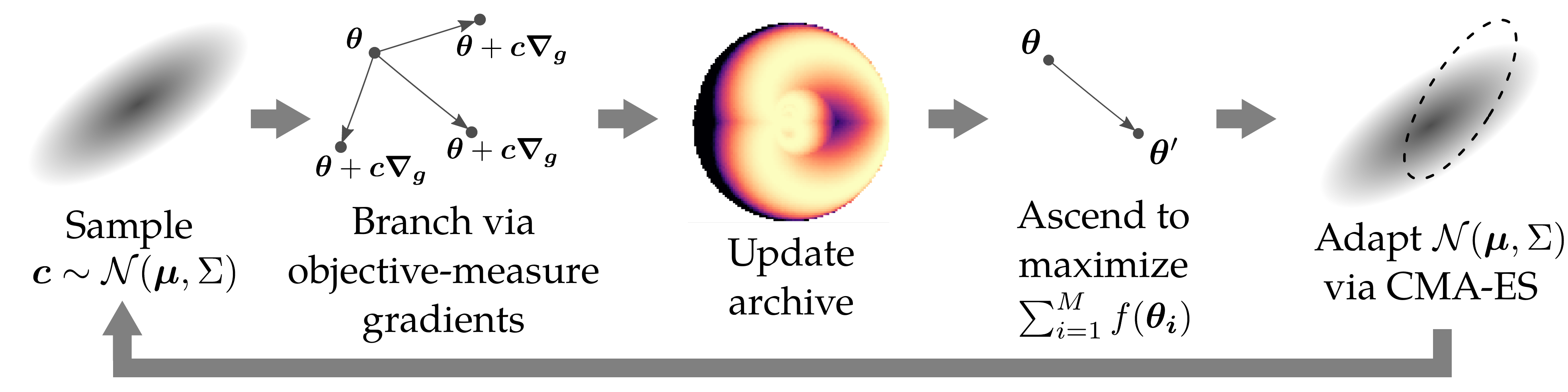}
\caption{An overview of the Covariance Matrix Adaptation MAP-Elites via a Gradient Arborescence (CMA-MEGA) algorithm. The algorithm leverages a gradient arborescence to branch in objective-measure space, while dynamically adapting the gradient steps to maximize a QD objective (Eq.~\ref{eq:objective}).}
\label{fig:front}
\end{figure}

Our work makes four main contributions. 1) We introduce and formalize the problem of differentiable quality diversity (DQD). 2) We propose two DQD algorithms: Objective and Measure Gradient \mbox{MAP-Elites} via a Gradient Arborescence (OMG-MEGA), an algorithm based on MAP-Elites~\cite{cully:nature15}, which branches based on the measures $m_i$ but ascends based on the objective function $f$;  and Covariance Matrix Adaptation MEGA (CMA-MEGA) which is based on the CMA-ME~\cite{fontaine2020covariance} algorithm, and which branches based on the objective-measure space but ascends based on maximizing the QD objective (Fig.~\ref{fig:front}). Both algorithms search directly in measure space and leverage the gradients of $f$ and $m_i$ to form efficient parameter space steps in $\bm{\theta}$. 3) We show in three different QD domains (the linear projection, the arm repertoire, and the latent space illumination (LSI) domains), that DQD algorithms significantly outperform state-of-the-art QD algorithms that treat the objective and measure functions as a black box. 4) We demonstrate how searching the latent space of a StyleGAN~\cite{karras2019style} in the LSI domain with CMA-MEGA results in a diverse range of high-quality images.

\section{Problem Definition } \label{sec:problem}
\noindent\textbf{Quality Diversity.} 
The quality diversity (QD) problem assumes an objective $f: \mathbb{R}^n \rightarrow \mathbb{R}$ in an \mbox{$n$-dimensional} continuous space $\mathbb{R}^n$ and $k$ measures $m_i: \mathbb{R}^n \rightarrow \mathbb{R}$ or, as a joint measure, $\bm{m}: \mathbb{R}^n \rightarrow \mathbb{R}^k$.  Let $S = \bm{m}(\mathbb{R}^n)$ be the measure space formed by the range of $m$. For each $\bm{s} \in S$ the QD objective is to find a solution $\bm{\theta} \in \mathbb{R}^n$ such that $\bm{m}(\bm{\theta}) = \bm{s}$ and $f(\bm{\theta})$ is maximized.

However, we note that $\mathbb{R}^k$ is continuous, and an algorithm solving the quality diversity problem would require infinite memory to store all solutions. Thus, QD algorithms in the MAP-Elites~\citep{mouret2015illuminating,cully:nature15} family approximate the problem by discretizing $S$ via a tessellation method. Let $T$ be the tessellation of $S$ into $M$ cells. We relax the QD objective to find a set of solutions $\bm{\theta_i}, i \in \{1,\ldots,M\}$, such that each $\bm{\theta_i}$ occupies one unique cell in $T$. The occupants  $\bm{\theta_i}$ of all $M$ cells  form an archive of solutions. Each solution $\bm{\theta_i}$ has a position in the archive $\bm{m}(\bm{\theta_i})$, corresponding to one out of $M$ cells, and an objective value $f(\bm{\theta_i})$.

 The objective of QD can be rewritten as follows, where the goal is to maximize the objective value for each cell in the archive:

\begin{equation}
  \max \sum_{i=1}^M f(\bm{\theta_i})
\label{eq:objective}
\end{equation}

\noindent\textbf{Differentiable Quality Diversity.} We define the differentiable quality diversity (DQD) problem, as a QD problem where both the objective $f$ and measures $m_i$ are \textit{first-order differentiable}.  

\section{Preliminaries} \label{sec:preliminaries}

We present several state-of-the-art derivative-free QD algorithms. Our proposed DQD algorithm MEGA builds upon ideas from these works, while introducing measure and objective gradients into the optimization process.

\noindent\textbf{MAP-Elites and MAP-Elites (line)}. 
MAP-Elites~\cite{cully:nature15,mouret2015illuminating} first tessellates the measure space $S$ into evenly-spaced grid cells. The upper and lower bounds for $\bm{m}$ are given as input to constrain $S$ to a finite region. MAP-Elites first samples solutions from a fixed distribution $\bm{\theta} \sim \mathcal{N}(\bm{0},I)$, and populates an initial archive after computing $f(\bm{\theta})$ and $\bm{m}(\bm{\theta})$. Each iteration of MAP-Elites selects $\lambda$ cells uniformly at random from the archive and perturbs each occupant $\bm{\theta_i}$ with fixed-variance $\sigma$ isotropic Gaussian noise: $\bm{\theta}' = \bm{\theta_i} + \sigma \mathcal{N}(\bm{0},I)$. Each new candidate solution $\bm{\theta'}$ is then evaluated and added to the archive if $\bm{\theta'}$ discovers a new cell or improves an existing cell. The algorithm continues to generate solutions for a specified number of iterations.

Later work introduced the Iso+LineDD operator~\citep{vassiliades:gecco18}. The Iso+LineDD operator samples two archive solutions $\bm{\theta_i}$ and $\bm{\theta_j$}, then blends a Gaussian perturbation with a noisy interpolation given hyperparameters $\sigma_1$ and $\sigma_2$: \mbox{$\bm{\theta'} = \bm{\theta_i} + \sigma_1 \mathcal{N}(\bm{0},I) +  \sigma_2 \mathcal{N}(\bm{0},1) (\bm{\theta_i} - \bm{\theta_j})$}. In this paper we refer to MAP-Elites with an Iso+LineDD operator as MAP-Elites (line).

\noindent\textbf{CMA-ME}. \label{subsec:CMA-ME} 
Covariance Matrix Adaptation MAP-Elites (\mbox{CMA-ME})~\citep{fontaine2020covariance} combines the archiving mechanisms of MAP-Elites with the adaptation mechanisms of CMA-ES~\cite{hansen:cma16}. While MAP-Elites creates new solutions by perturbing existing solutions with fixed-variance Gaussian noise, \mbox{CMA-ME} maintains a full-rank Gaussian distribution $\mathcal{N}(\bm{\mu},\Sigma)$ in parameter space $\mathbb{R}^n$. Each iteration of CMA-ME samples $\lambda$ candidate solutions $\bm{\theta_i} \sim \mathcal{N}(\bm{\mu},\Sigma)$, evaluates each solution, and updates the archive based on the following rule: if there is a previous occupant 
$\bm{\theta_p}$  at the same cell, we compute  $\Delta_i = f(\bm{\theta_i}) - f(\bm{\theta_p})$, otherwise if the cell is empty we compute  $\Delta_i = f(\bm{\theta_i})$. We then rank the sampled solutions by increasing improvement $\Delta_i$, with an extra criteria that candidates discovering new cells are ranked higher than candidates that improve existing cells. We then update $\mathcal{N}(\bm{\mu},\Sigma)$ with the standard CMA-ES update rules based on the improvement ranking. CMA-ME restarts when all $\lambda$ solutions fail to change the archive. On a restart we reset the Gaussian $\mathcal{N}(\bm{\theta_i},I)$, where $\bm{\theta_i}$ is an archive solution chosen uniformly at random, and all internal CMA-ES parameters. In the supplemental material, we derive, for the first time, a natural gradient interpretation of CMA-ME's improvement ranking mechanism, based on previous theoretical work on CMA-ES~\cite{akimoto2010bidirectional}.

\section{Algorithms} \label{sec:algorithms}

We present two variants of our proposed MEGA algorithm:   OMG-MEGA and CMA-MEGA. We form each variant by adapting the concept of a gradient arborescence to the MAP-Elites and CMA-ME algorithms, respectively. Finally, we introduce two additional baseline algorithms, OG-MAP-Elites and OG-MAP-Elites (line), which operate only on the gradients of the objective.


\noindent\textbf{OMG-MEGA.} We first derive the Objective and Measure Gradient MAP-Elites via Gradient Arborescence (OMG-MEGA) algorithm from MAP-Elites. 

First, we observe how gradient information could benefit a QD algorithm. Note that the QD objective is to explore the measure space, while maximizing the objective function $f$. We observe that maximizing a linear combination of measures : $\sum_{j=1}^{k} c_j m_j(\bm{\theta})$, where $\bm{c}$ is a $k$-dimensional vector of coefficients, enables movement in a $k$-dimensional measure space. Adding the objective function $f$  to the linear sum enables movement in an objective-measure space. Maximizing $g$ with a positive coefficient of $f$ results in steps that increasing $f$, while the direction of movement in the measure space is determined by the sign and magnitude of the coefficients $c_j$.

\begin{equation}
g(\bm{\theta}) = \abs{c_0} f(\bm{\theta}) + \sum_{j=1}^{k} c_j m_j(\bm{\theta})
\end{equation}

We can then derive a direction function that perturbs a given solution $\bm{\theta}$ based on the gradient of our linear combination $g$: \mbox{$\bm{\nabla}g(\bm{\theta}) = \abs{c_0} \bm{\nabla} f(\bm{\theta}) + \sum_{j=1}^{k} c_j \bm{\nabla} m_j(\bm{\theta})$ }. We incorporate the direction function $\bm{\nabla} g$ to derive a gradient-based MAP-Elites variation operator.

We observe that MAP-Elites samples a cell $\bm{\theta_i}$ and perturbs the occupant with Gaussian noise: $\bm{\theta'} = \bm{\theta_i} + \sigma \mathcal{N}(\bm{0}, I)$. Instead, we sample coefficents $\bm{c} \sim \mathcal{N}(\bm{0}, \sigma_g I)$ and step: 

\begin{equation}
\bm{\theta'} = \bm{\theta_i} +  \mid c_0 \mid \bm{\nabla} f(\bm{\theta_i}) + \sum_{j=1}^{k} c_j \bm{\nabla}  m_j(\bm{\theta_i})
\label{eq:update}
\end{equation}

The value $\sigma_g$ acts as a learning rate for the gradient step, because it controls the scale of the coefficients $c\sim \mathcal{N}(0, \sigma_g I)$. To balance the contribution of each function, we normalize all gradients. In the supplemental material, we further justify gradient normalization and provide an empirical ablation study. Other than our new gradient-based operator, OMG-MEGA is identical to MAP-Elites.

\noindent\textbf{CMA-MEGA.} Next, we derive the Covariance Matrix Adaptation MAP-Elites via a Gradient Arborescence (CMA-MEGA) algorithm from CMA-ME. Fig.~\ref{fig:front} shows an overview of the algorithm.

First, we note that we sample $\bm{c}$ in OMG-MEGA from a fixed-variance Gaussian. However, it would be beneficial to select $\bm{c}$ based on how $\bm{c}$, and the subsequent gradient step on $\bm{\theta}$, improve the QD objective defined in equation~\ref{eq:objective}.

We frame the selection of $\bm{c}$ as an optimization problem with the objective of maximizing the QD objective (Eq.~\ref{eq:objective}). We model a distribution of coefficients $\bm{c}$ as a $k+1$-dimensional Gaussian $\mathcal{N}(\bm{\mu}, \Sigma)$. Given a $\bm{\theta}$, we can sample  $\bm{c} \sim \mathcal{N}(\bm{\mu}, \Sigma)$, compute $\bm{\theta'}$ via Eq.~\ref{eq:update}, and adapt $\mathcal{N}(\bm{\mu}, \Sigma)$ towards the direction of maximum increase of the QD objective (see Eq.~\ref{eq:objective}).

We follow an evolution strategy approach to model and dynamically adapt the sampling distribution of coefficients $\mathcal{N}(\bm{\mu}, \Sigma)$.  We sample a population of $\lambda$ coefficients from $\bm{c_i} \sim \mathcal{N}(\bm{\mu}, \Sigma)$ and generate $\lambda$ solutions $\bm{\theta_i}$. We then compute $\Delta_i$ from CMA-ME's improvement ranking for each candidate solution $\bm{\theta_i}$. By updating $\mathcal{N}(\bm{\mu}, \Sigma)$ with CMA-ES update rules for the ranking $\Delta_i$, we dynamically adapt the distribution of coefficients $\bm{c}$ to maximize the QD objective.

Algorithm~\ref{alg:cma-mega} shows the pseudocode for CMA-MEGA. In line~\ref{lst:initeval} we evaluate the current solution and compute an objective value $f$, a vector of measure values $\bm{m}$, and gradient vectors. As we dynamically adapt the coefficients $\bm{c}$, we normalize the objective and measure gradients (line~\ref{lst:norm}) for stability. Because the measure space is tessellated, the measures $\bm{m}$ place solution $\bm{\theta}$ into one of the $M$ cells in the archive. We then add the solution to the archive (line~\ref{lst:archiveupdate}), if the solution discovers an empty cell in the archive, or if it improves an existing cell, identically to MAP-Elites.

We then use the gradient information to compute a step that maximizes improvement of the archive. In lines~\ref{lst:startforloop}-\ref{lst:endforloop}, we sample a population  of  $\lambda$ coefficients from a multi-variate Gaussian retained by CMA-ES, and take a gradient step for each sample. We evaluate each sampled solution $\bm{\theta_i}'$, and compute the improvement $\Delta_i$ (line~\ref{lst:imprv}). As in CMA-ME, we specify $\Delta_i$ as the difference in the objective value between the sampled solution $\bm{\theta}_i$ and the existing solution, if one exists, or as the absolute objective value of the sampled solution if $\bm{\theta_i}$ belongs to an empty cell. 

In line~\ref{lst:ranking}, we rank the sampled gradients $\bm{\nabla}_i$ based on their respective improvements. As in \mbox{CMA-ME}, we prioritize exploration of the archive by ranking first by their objective values all samples that discover new cells, and subsequently all samples that improve existing cells by their difference in improvement. We then compute an ascending gradient step as a linear combination of gradients (line~\ref{lst:gradient_linear_comb}), following the recombination weights $w_i$ from CMA-ES~\citep{hansen:cma16} based on the computed improvement ranking. These weights correspond to the log-likelihood probabilities of the samples in the natural gradient interpretation of CMA-ES~\citep{akimoto2010bidirectional}.



In line~\ref{lst:adaptation}, CMA-ES adapts the multi-variate Gaussian $\mathcal{N}(\bm{\mu}, \Sigma)$, as well as internal search parameters $\bm{p}$, from the improvement ranking of the coefficients. In the supplemental material, we provide a natural gradient interpretation of the improvement ranking rules of CMA-MEGA, where we show that the coefficient distribution of CMA-MEGA approximates natural gradient steps of maximizing a modified QD objective.


\noindent\textbf{CMA-MEGA (Adam).} We add an Adam-based variant of CMA-MEGA, where we replace line~\ref{lst:gradient_step} with an Adam gradient optimization step~\cite{kingma2014adam}.

\noindent\textbf{OG-MAP-Elites.} 
To show the importance of taking gradient steps in the measure space, as opposed to only taking gradient steps in objective space and directly perturbing the parameters, we derive two variants of MAP-Elites as a baseline that draw insights from the recently proposed Policy Gradient Assisted MAP-Elites (PGA-ME) algorithm~\citep{nilsson2021policy}. PGA-ME combines the Iso+LineDD operator~\citep{vassiliades:gecco18} with a policy gradient operator only on the objective. Similarly, our proposed Objective-Gradient MAP-Elites (OG-MAP-Elites) algorithm combines an objective gradient step with a MAP-Elites style perturbation operator. Each iteration of OG-MAP-Elites samples $\lambda$ solutions $\bm{\theta_i}$ from the archive. Each $\bm{\theta_i}$ is perturbed with Gaussian noise to form a new candidate solution $\bm{\theta_i}' = \bm{\theta_i} + \sigma \mathcal{N}(\bm{0},I)$. OG-MAP-Elites evaluates the solution and updates the archive, exactly as in MAP-Elites. However, OG-MAP-Elites takes one additional step: for each $\bm{\theta_i}'$, the algorithm computes $\bm{\nabla} f(\bm{\theta_i}')$, forms a new solution  $\bm{\theta_i}'' = \bm{\theta_i}' + \eta \bm{\nabla} f(\bm{\theta_i}')$ with an objective gradient step, and evaluates $\bm{\theta_i}''$.  Finally, we update the archive with all solutions $\bm{\theta_i}'$ and $\bm{\theta_i}''$. 

\noindent\textbf{OG-MAP-Elites (line).} Our second baseline, OG-MAP-Elites (line) replaces the Gaussian operator with the Iso+LineDD operator~\citep{vassiliades:gecco18}: \mbox{$\bm{\theta'} = \bm{\theta_i} + \sigma_1 \mathcal{N}(\bm{0},I) +  \sigma_2 \mathcal{N}(\bm{0},1) (\bm{\theta_i} - \bm{\theta_j})$}. We consider \mbox{OG-MAP-Elites (line)} a DQD variant of PGA-ME. However, PGA-ME was designed as a reinforcement learning (RL) algorithm, thus many of the advantages gained in RL settings are lost in OG-MAP-Elites (line). We provide a detailed discussion and ablations in the supplemental material.

\begin{algorithm}[t!]
\SetAlgoLined
\caption{Covariance Matrix Adaptation MAP-Elites via a Gradient Aborescence (CMA-MEGA)}
\label{alg:cma-mega}
\SetKwInOut{Input}{input}
\SetKwInOut{Result}{result}
\SetKwProg{CMAMEGA}{CMA-MEGA}{}{}
\DontPrintSemicolon
\CMAMEGA{$(evaluate, \bm{\theta_0}, N, \lambda, \eta, \sigma_g)$}
{
\Input{An evaluation function $evaluate$ which computes the objective, the measures, gradients of the objective and measures, an initial solution $\bm{\theta_0}$, a desired number of iterations $N$, a branching population size $\lambda$, a learning rate $\eta$, and an initial step size for CMA-ES $\sigma_g$.}
\Result{Generate $N (\lambda + 1)$ solutions storing elites in an archive $A$.}

\BlankLine
Initialize solution parameters $\bm{\theta}$ to $\bm{\theta_0}$, CMA-ES parameters $\bm{\mu}=\bm{0}$, $\Sigma=\sigma_g I$, and $\bm{p}$, where we let $\bm{p}$ be the CMA-ES internal parameters.

\For{$iter\leftarrow 1$ \KwTo $N$}{
$f, \bm{\nabla}_{f}, \bm{m}, \bm{\nabla_m} \gets  \mbox{evaluate}(\bm{\theta})$\; \label{lst:initeval}

$\bm{\nabla}_{f} \gets \mbox{normalize}(\bm{\nabla}_{f}),\bm{\nabla}_{m} \gets \mbox{normalize}(\bm{\nabla}_{m})$\;\label{lst:norm}


$\mbox{update\_archive}(\bm{\theta},f,\bm{m})$\; \label{lst:archiveupdate} 

\For{$i\leftarrow 1$ \KwTo $\lambda$}{ \label{lst:startforloop} 
  $\bm{c} \sim \mathcal{N}(\bm{\mu},\Sigma)$\;

  $\bm{\nabla}_i \gets c_0 \bm{\nabla}_f + \sum_{j=1}^{k} c_j \bm{\nabla}_{m_j}$\;\label{lst:line2}

  $\bm{\theta_i}' \gets \bm{\theta} + \bm{\nabla}_i$\;

 $f', *, \bm{m}',* \gets \mbox{evaluate}(\bm{\theta_i}')$\;

  $\Delta_i \gets \mbox{update\_archive}(\bm{\theta_i}',f',\bm{m}')$\;\label{lst:imprv}

}  \label{lst:endforloop}
rank $\bm{\nabla}_i$ by $\Delta_i$ \; \label{lst:ranking}

$\bm{\nabla}_{\textrm{step}} 
\gets \sum_{i=1}^{\lambda}w_i  \bm{\nabla}_{\textrm{rank[i]}}$\; \label{lst:gradient_linear_comb}

$\bm{\theta} \gets \bm{\theta} +  \eta \bm{\nabla}_{\textrm{step}}$\; \label{lst:gradient_step}

Adapt CMA-ES parameters $\bm{\mu},\Sigma,\bm{p}$ based on improvement ranking $\Delta_i$\label{lst:adaptation}

\If{\mbox{there is no change in the archive}}{
Restart CMA-ES with $\bm{\mu}=0, \Sigma = \sigma_g I$.

Set $\bm{\theta}$ to a randomly selected existing cell $\bm{\theta_i}$  from the archive

}
}

}

\end{algorithm}

\section{Domains}

DQD requires differentiable objective and measures, thus we select benchmark domains from previous work in the QD literature where we can compute the gradients of the objective and measure functions.

\noindent\textbf{Linear Projection.} To show the importance of adaptation mechanisms in QD, the \mbox{CMA-ME} paper~\citep{fontaine2020covariance} introduced a simple domain, where reaching the extremes of the measures is challenging for non-adaptive QD algorithms. The domain forms each measure $m_i$ by a linear projection from $\mathbb{R} ^ n$ to $\mathbb{R}$, while bounding the contribution of each component $\bm{\theta}_i$ to the range $[-5.12, 5.12]$.

We note that uniformly sampling from a hypercube in $\mathbb{R}^n$  results in a narrow distribution of the linear projection in $\mathbb{R}$~\citep{fontaine2020covariance,johnson1995continuous}. Increasing the number of parameters $n$ makes the problem of covering the measure space more challenging, because to reach an extremum $m_i(\bm{\theta}) = \pm 5.12n$, all components must equal the extremum: $\bm{\theta[i]} = \pm 5.12$.

We select this domain as a benchmark to highlight the need for adaptive gradient coefficients for CMA-MEGA as opposed to constant coefficients for OMG-MEGA, because reaching the edges of the measure space requires dynamically shrinking the gradient steps.

As a QD domain, the domain must provide an objective. The CMA-ME study~\cite{fontaine2020covariance} introduces two variants of the linear projection domain with an objective based on the sphere and Rastrigin functions from the continuous black-box optimization set of benchmarks~\citep{hansen:arxiv16,hansen:gecco10}. We optimize an $n=1000$ unbounded parameter space $\mathbb{R}^n$. We provide more detail in the supplemental material.

\noindent\textbf{Arm Repertoire.} We select the robotic arm repertoire domain from previous work~\cite{cully:nature15,vassiliades:gecco18}. The goal in this domain is to find an inverse kinematics (IK) solution for each reachable position of the end-effector of a planar robotic arm with revolute joints. The objective $f$ of each solution is to minimize the variance of the joint angles, while the measure functions are the positions of the end effector in the $x$ and $y$-axis, computed with the forward kinematics of the planar arm~\cite{murray2017mathematical}. We selected a 1000-DOF robotic arm.

\noindent\textbf{Latent Space Illumination.} Previous work~\citep{fontaine2020illuminating} introduced the problem of exploring the latent space of a generative model directly with a QD algorithm. The authors named the problem latent space illumination (LSI). As the original LSI work evaluated non-differentiable objectives and measures, we create a new benchmark for the differentiable LSI problem by generating images with StyleGAN~\citep{karras2019style} and leveraging CLIP~\citep{radford2021learning} to create differentiable objective and measure functions. We adopt the StyleGAN+CLIP~\cite{styleclip} pipeline, where StyleGAN-generated images are passed to CLIP, which in turn evaluates how well the generated image matches a given text prompt. We form the prompt ``Elon Musk with short hair.'' as the objective and for the measures we form the prompts ``A person with red hair.'' and ``A man with blue eyes.''. The goal of DQD becomes generating faces similar to Elon Musk with short hair, but varying with respect to hair and eye color.

\begin{table*}[t]
\centering
\resizebox{1.0\linewidth}{!}{
\begin{tabular}{l|rr|rr|rr|rr}
\hline
             & \multicolumn{2}{l|}{LP (sphere)}     & \multicolumn{2}{l|}{LP (Rastrigin)}          & \multicolumn{2}{l|}{Arm Repertoire} & \multicolumn{2}{l|}{LSI} \ \\ 
    \toprule
Algorithm     & QD-score & Coverage  & QD-score & Coverage   & QD-score & Coverage  & QD-score & Coverage \\
    \midrule
MAP-Elites  &  1.04 & 1.17\%  &  1.18 & 1.72\% & 1.97 & 8.06\% & 13.88 & 23.15\%\\
MAP-Elites (line)   & 12.21 & 14.32\%  & 8.12 & 11.79\% & 33.51 & 35.79\%  & 16.54 & 25.73\% \\
CMA-ME    & 1.08 & 1.21\% & 1.21 & 1.76\% &  55.98 & 56.95\%  &  18.96 & 26.18\% \\
 OG-MAP-Elites  & 1.52 & 1.67\% & 0.83 & 1.26\% &  57.17 & 58.08\%  &  N/A & N/A\\
 OG-MAP-Elites (line) & 15.01 & 17.41\% & 6.10 & 8.85\%& 59.66 & 60.28\% & N/A & N/A \\ 
OMG-MEGA   & 71.58 & 92.09\% & 55.90 & 77.00\% &  44.12 & 44.13\% &  N/A & N/A \\
CMA-MEGA   & 75.29 & \textbf{100.00\%} & 62.54 & \textbf{100.00\%}  &  \textbf{74.18}& \textbf{74.18\%}  &  5.36& 8.61\% \\
CMA-MEGA (Adam)   & \textbf{75.30} & \textbf{100.00\%} & \textbf{62.58} & \textbf{100.00\%} & 73.82 & 73.82\%  & \textbf{21.82} & \textbf{30.73\%} \\

  \bottomrule
\end{tabular}
}
\caption{Mean QD-score and coverage values after 10,000 iterations for each algorithm per domain.}
\label{tab:results}
\end{table*}

\section{Experiments} \label{sec:experiments}
We conduct experiments to assess the performance of the MEGA variants. In addition to our \mbox{OG-MAP-Elites} baselines, which we propose in section~\ref{sec:algorithms}, we compare the MEGA variants with the state-of-the-art QD algorithms presented in section~\ref{sec:preliminaries}. We implemented MEGA and OG-MAP-Elites variants in the Pyribs~\citep{pyribs} QD library and compare against the existing Pyribs implementations of  MAP-Elites, MAP-Elites (line), and CMA-ME.

\subsection{Experiment Design} \label{subsec:design}
\noindent\textbf{Independent Variables.} 
We follow a between-groups design, where the independent variables are the algorithm and the domain (linear projection, arm repertoire, and LSI). We did not run \mbox{OMG-MEGA} and OG-MAP-Elites in the LSI domain; while CMA-MEGA computes the $f$ and $m_i$ gradients once per iteration (line~\ref{lst:initeval} in Algorithm~\ref{alg:cma-mega}), OMG-MEGA and OG-MAP-Elites compute the $f$ and $m_i$ gradients for every sampled solution, making their execution cost-prohibitive for the LSI domain.

\noindent\textbf{Dependent Variables.} We measure both the diversity and the quality of the solutions returned by each algorithm. These are combined by the QD-score metric~\cite{pugh2015confronting}, which is defined as the sum of $f$ values of all cells in the archive (Eq.~\ref{eq:objective}). To make the QD-score invariant with respect to the resolution of the archive, we normalize QD-score by the archive size (the total number of cells from the tessellation of the measure space). As an additional metric of diversity we compute the coverage as the number of occupied cells in the archive divided by the total number of cells. We run each algorithm for 20 trials in the linear projection and arm repertoire domains, and for 5 trials in the LSI domain, resulting in a total of 445 trials.



\begin{figure*}[t!]
\includegraphics[width=1.0\columnwidth]{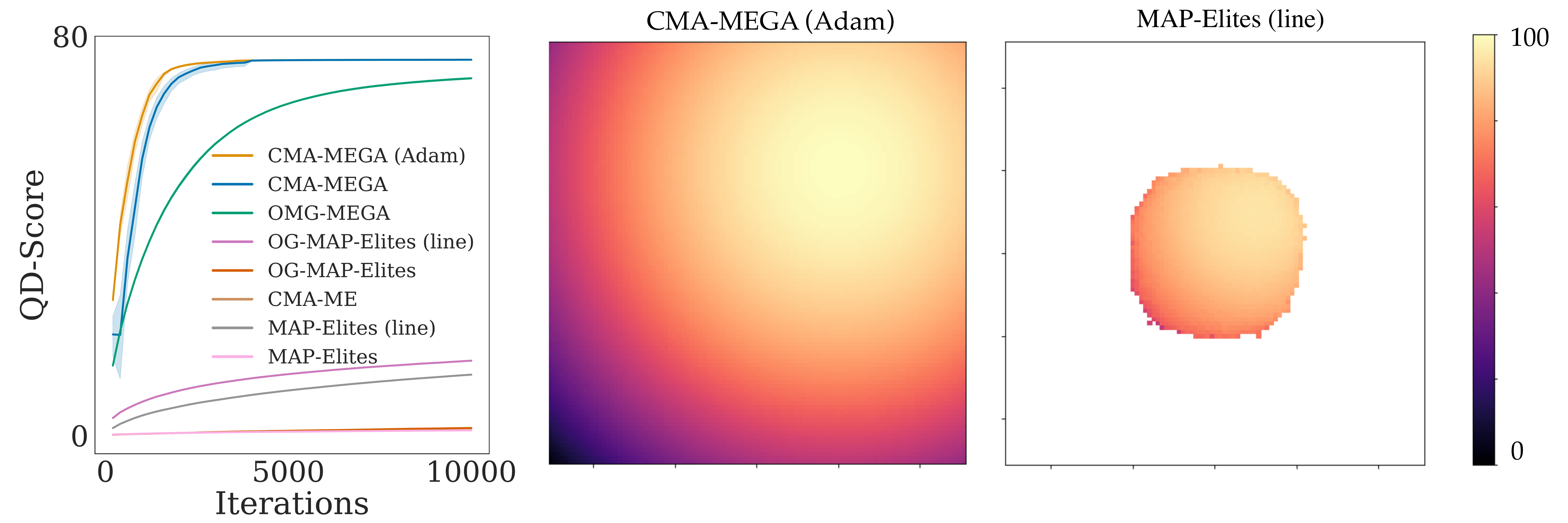}
\includegraphics[width=1.0\columnwidth]{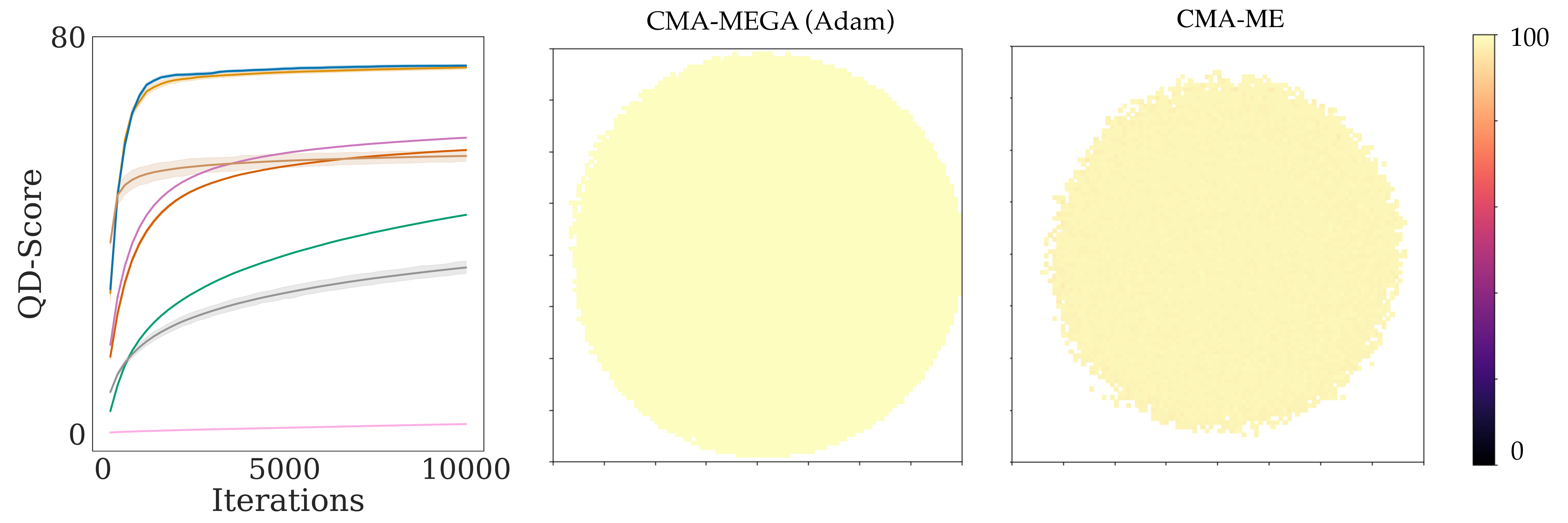}
\includegraphics[width=1.0\columnwidth]{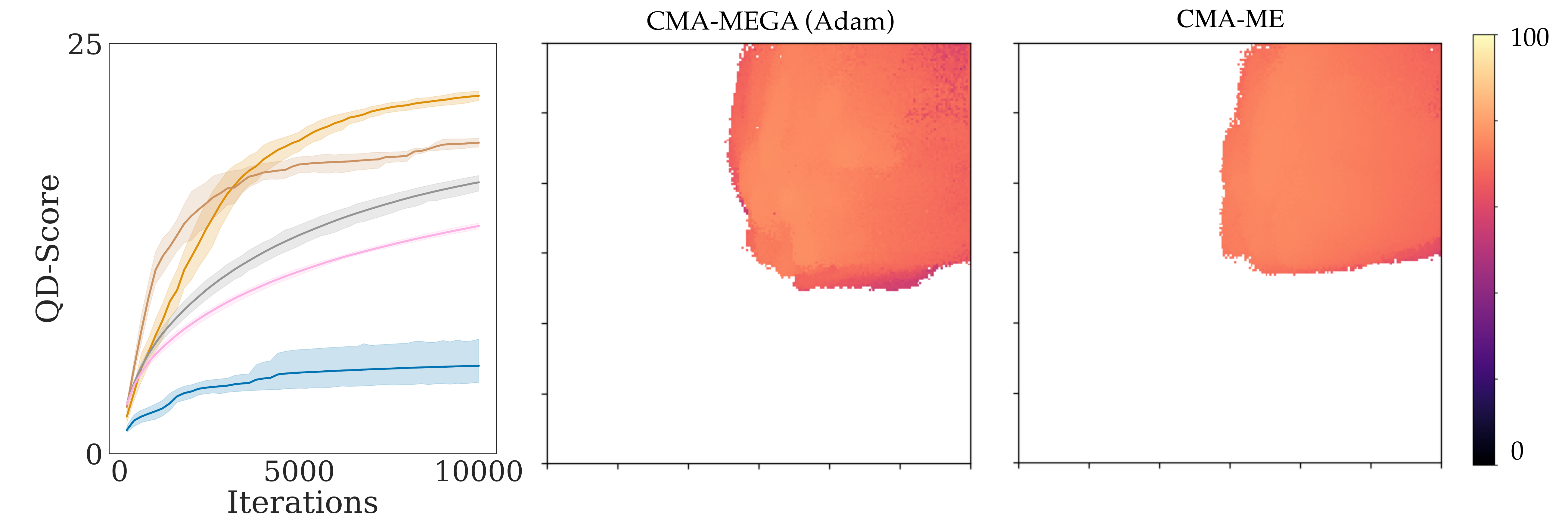}
\caption{QD-Score plot with 95$\%$ confidence intervals and heatmaps of generated archives by CMA-MEGA (Adam) and the strongest derivative-free competitor for the linear projection sphere (top), arm repertoire (middle), and latent space illumination (bottom) domains. }
\label{fig:summary}
\end{figure*}

\subsection{Analysis}

Table~\ref{tab:results} shows the metrics of all the algorithms, averaged over 20 trials for the benchmark domains and over 5 trials for the LSI domain. We conducted a two-way ANOVA to examine the effect of algorithm and domain (linear projection (sphere), linear projection (Rastrigin), arm repertoire) on the QD-Score. There was a statistically significant interaction between the search algorithm and the domain $(F(14,456) = 7328.18, p < 0.001$). Simple main effects analysis with Bonferroni corrections showed that \mbox{CMA-MEGA} and \mbox{OMG-MEGA} performed significantly better than each of the baselines in the sphere and Rastrigin domains, with CMA-MEGA significantly outperforming OMG-MEGA. CMA-MEGA also outperformed all the other algorithms in the arm repertoire domain.

We additionally conducted a one-way ANOVA to examine the effect of algorithm on the LSI domain. There was a statistically significant difference between groups ($F(4,20)=260.64, p< 0.001$). Post-hoc pairwise comparisons with Bonferroni corrections showed that CMA-MEGA (Adam) significantly outperformed all other algorithms, while CMA-MEGA without the Adam implementation had the worst performance.

Both OMG-MEGA and CMA-MEGA variants perform well in the linear projection domain, where the objective and measure functions are additively separable, and the partial derivatives with respect to each parameter independently capture the steepest change of each function. We observe that \mbox{OG-MAP-Elites} performs poorly in this domain. Analysis shows that the algorithm finds a nearly perfect best solution for the sphere objective, but it interleaves following the gradient of the objective with exploring the archive as in standard MAP-Elites, resulting in smaller coverage of the archive. 

In the arm domain, OMG-MEGA manages to reach the extremes of the measure space, but the algorithm fails to fill in nearby cells. The OG-MAP-Elites variants perform significantly better than \mbox{OMG-MEGA}, because the top-performing solutions in this domain tend to be concentrated in an ``elite hypervolume''~\cite{vassiliades:gecco18}; moving towards the gradient of the objective finds top-performing cells, while applying isotropic perturbations to these cells fills in nearby regions in the archive. CMA-MEGA variants retain the best performance in this domain. Fig.~\ref{fig:front} shows a high-precision view of the \mbox{CMA-MEGA (Adam)} archive for the arm repertoire domain.

\begin{figure}[!t]
\centering
\includegraphics[width=1.0\linewidth]{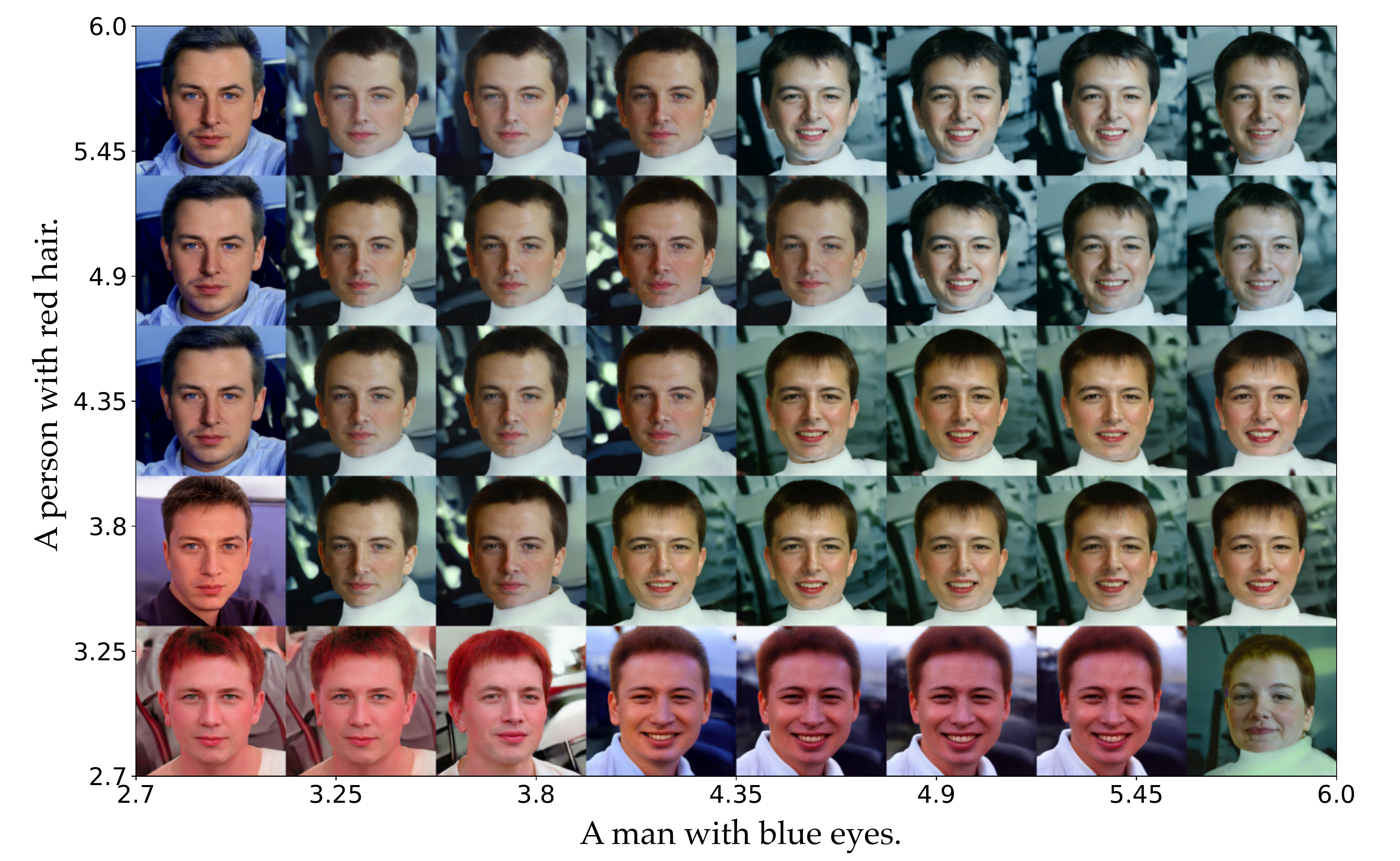}
\caption{Result of latent space illumination for the objective prompt ``Elon Musk with short hair.'' and for the measure prompts ``A person with red hair.'' and ``A man with blue eyes.''. The axes values indicate the score returned by the CLIP model, where lower score indicates a better match.}

\label{fig:collage}
\end{figure}

We did not observe a large difference between the CMA-MEGA (Adam) and our gradient descent implementation in the first two benchmark domains, where the curvature of the search space is well-conditioned. On the other hand, in the LSI domain CMA-MEGA without the Adam implementation performed poorly. We conjecture that this is caused by the conditioning of the mapping from the latent space of the StyleGAN to the CLIP score.

 Fig.~\ref{fig:summary} shows the QD-score values for increasing number of iterations for each of the tested algorithms, with 95\% confidence intervals. The figure also presents heatmaps of the CMA-MEGA (Adam) and the generated archive of the strongest QD competitor for each of the three domains. We provide generated archives of all algorithms in the supplemental material.
 
 We visualize the top performing solutions in the LSI domain by uniformly sampling solutions from the archive of CMA-MEGA (Adam) and showing the generated faces in Fig.~\ref{fig:collage}. We observe that as we move from the top right to the bottom left, the features matching the captions ``a man with blue eyes'' and ``a person with red hair'' become more prevalent. We note that these solutions were generated from a single run of CMA-MEGA (Adam) for 10,000 iterations.

Overall, these results show that using the gradient information in quality diversity optimization results in significant benefits to search efficiency, but adapting the gradient coefficients with CMA-ES is critical in achieving top performance. 

\section{Related Work} 

\noindent\textbf{Quality Diversity}. The precursor to QD algorithms~\citep{pugh2016quality} originated with diversity-driven algorithms as a branch of evolutionary computation. Novelty search~\citep{lehman2011abandoning}, which maintains an archive of diverse solutions, ensures diversity though a provided metric function and was the first diversity-driven algorithm. Later, objectives were introduced as a quality metric resulting in the first QD algorithms: Novelty Search with Local Competition (NSLC)~\citep{lehman2011evolving} and MAP-Elites~\citep{cully:nature15,mouret2015illuminating}. Since their inception, many works have improved the archives~\citep{fontaine2019mapping,vassiliades2016using,smith2016rapid}, the variation operators~\citep{vassiliades:gecco18,fontaine2020covariance,conti2017improving,nordmoen2018dynamic}, and the selection mechanisms~\citep{cully2017quality,sfikas2021monte} of both NSLC and MAP-Elites. While the original QD algorithms were based on genetic algorithms, algorithms based on other derivative-free approaches such as evolution strategies~\citep{fontaine2020covariance,colas2020scaling,nordmoen2018dynamic,conti2017improving} and Bayesian optimization~\citep{kent2020bop} have recently emerged.

%
%

%

Being stochastic derivative-free optimizers~\citep{chatzilygeroudis2020quality}, QD algorithms are frequently applied to reinforcement learning (RL) problems~\citep{parker2020effective,arulkumaran2019alphastar,ecoffet2021first} as derivative information must be estimated in RL. Naturally, approaches combining QD and RL have started to emerge~\cite{nilsson2021policy,cideron2020qd}. Unlike DQD, these approaches \emph{estimate} the gradient of the reward function, and in the case of QD-RL a novelty function, in action space and backpropagate this gradient through a neural network. Our proposed DQD problem differs by leveraging \textit{provided} -- rather than approximated -- gradients of the objective and measure functions.

%

Several works have proposed model-based QD algorithms. For example, the DDE-Elites algorithm~\citep{gaier2020automating} dynamically trains a variational autoencoder (VAE) on the MAP-Elites archive, then leverages the latent space of this VAE by interpolating between archive solutions in latent space as a type of crossover operator. DDE-Elites learns a manifold of the archive data as a representation, regularized by the VAE's loss function, to solve downstream optimization tasks efficiently in this learned representation. The PoMS algorithm~\citep{rakicevic2021policy} builds upon DDE-Elites by learning an \textit{explicit} manifold of the archive data via an autoencoder. To overcome distortions introduced by an explicit manifold mapping, the authors introduce a covariance perturbation operator based on the Jacobian of the decoder network. These works differ from DQD by dynamically constructing a learned representation of the search space instead of leveraging the objective and measure gradients directly.


\noindent\textbf{Latent Space Exploration}. Several works have proposed a variety of methods for directly exploring the latent space of generative models. Methods on GANs include interpolation~\citep{upchurch2017deep}, gradient descent~\citep{bojanowski2017optimizing}, importance sampling~\citep{white2016sampling}, and latent space walks~\citep{jahanian2019steerability}. Derivative-free optimization methods mostly consist of latent variable evolution (LVE)~\citep{bontrager2018deepmasterprints,galatolo2021generating}, the method of optimizing latent space with an evolutionary algorithm. LVE was later applied to generating Mario levels~\citep{volz2018evolving} with targeted gameplay characteristics. Later work~\citep{fontaine2020illuminating} proposed latent space illumination (LSI), the problem of exploring the latent space of a generative model with a QD algorithm. The method has only been applied to procedurally generating video game levels~\citep{fontaine2020illuminating,steckel2021illuminating,schrum2020cppn2gan,fontaine2021importance} and generating MNIST digits~\citep{pyribs_lsi_mnist}. Follow-up work explored LSI on VAEs~\citep{sarkar2021generating}. Our work improves LSI on domains where gradient information on the objective and measures is available with respect to model output.

%
%
%
%
%
%
%
%
%
%
%

\section{Societal Impacts} \label{sec:impacts}

By proposing gradient-based analogs to derivative-free QD methods, we hope to expand the potential applications of QD research and bring the ideas of the growing QD community to a wider machine learning audience. We are excited about future mixing of ideas between QD, generative modeling, and other machine learning subfields.

In the same way that gradient descent is used to synthesize super-resolution images~\cite{menon2020pulse}, our method can be used in the same context, which would raise ethical considerations due to potential biases present in the trained model~\cite{buolamwini2018gender}. On the other hand, we hypothesize that thoughtful selection of the measure functions may help counterbalance this issue, since we can explicitly specify the measures that ensure diversity over the collection of generated outputs. For example, a model may be \emph{capable} of generating a certain type of face, but the latent space may be \emph{organized} in a way which biases a gradient descent on the latent space away from a specific distribution of faces. If the kind of diversity required is differentiably measurable, then DQD could potentially help resolve which aspect of the generative model, i.e., the structure of the latent space or the representational capabilities of the model, is contributing to the bias.

Finally, we recognize the possibility of using this technology for malicious purposes, including generation of fake images (``DeepFakes''), and we highlight the utility of studies that help identify DeepFake models~\cite{guera2018deepfake}.

\section{Limitations and Future Work} \label{sec:discussion}

Quality diversity (QD) is a rapidly emerging field~\cite{chatzilygeroudis2020quality} with applications including procedural content generation~\cite{gravina2019procedural}, damage recovery in robotics~\cite{cully:nature15,mouret2015illuminating}, efficient aerodynamic shape design~\citep{gaier2018data}, and scenario generation in human-robot interaction~\cite{fontaine2020quality,fontaine2021importance}. We have introduced differentiable quality diversity (DQD), a special case of QD, where measure and objective functions are differentiable, and have shown how a gradient arborescence results in significant improvements in search efficiency. 

As both MEGA variants are only first order differentiable optimizers, we expect them to have difficulty on highly ill-conditioned optimization problems. CMA-ES, as an approximate second order optimizer, retains a full-rank covariance matrix that approximates curvature information and is known to outperform quasi-Newton methods on highly ill-conditioned problems~\cite{glasmachers2020hessian}. \mbox{CMA-ME} likely inherits these properties by leveraging the \mbox{CMA-ES} adaptation mechanisms and we expect it to have an advantage on ill-conditioned objective and measure functions.

While we found CMA-MEGA to be fairly robust to hyperparameter changes in the first two benchmark domains (linear projection, arm repertoire), small changes of the hyperparameters in the LSI domain led CMA-MEGA, as well as all the QD baselines, to stray too far from the mean of the latent space, which resulted in many artifacts and unrealistic images. One way to address this limitation is to constrain the search region to a hypersphere of radius $\sqrt{d}$, where $d$ is the dimensionality of the latent space, as done in previous work~\citep{menon2020pulse}. 

While CLIP achieves state-of-the-art performance in classifying images based on visual concepts, the model does not measure abstract concepts. Ideally, we would like to specify ``age'' as a measure function and obtain quantitative estimates of age given an image of a person. We believe that the proposed work on the LSI domain  will encourage future research on this topic, which we would in turn be able to integrate with DQD implementations to generate diverse, high quality content.

Many problems, currently modelled as optimization problems, may be fruitfully redefined as QD problems, including the training of deep neural networks. Our belief stems from recent works~\citep{ruchte2021efficient, liu2021stochastic}, which reformulated deep learning as a multi-objective optimization problem. However, QD algorithms struggle with high-variance stochastic objectives and measures~\citep{justesen2019map, flageat2020fast}, which naturally conflicts with minibatch training in stochastic gradient descent~\citep{bottou2012stochastic}. These challenges need to be addressed before DQD training of deep neural networks becomes tractable.

\begin{ack}

We would like to thank the anonymous reviewers fNzE, 4FYu, T1rp, and ghwT for their detailed feedback, thorough comments, and corrections throughout the review process that helped us improve the quality of the paper. We thank the area chair and senior area chair for their assessment, paper recommendation, and kind words about the paper. We thank Lisa B. Soros for her feedback on a preliminary version of this work and Varun Bhatt for his assistance running additional experiments for the final version of the paper.

This work was partially supported by the National Science Foundation NRI (\# 2024936) and the Alpha Foundation (\# AFC820-68). 


\end{ack}


{
\small
\bibliographystyle{plainnat}
\bibliography{references}
}

\appendix

\newpage

\section*{Appendix}
\section{Hyperparameter Selection}
For the arm and linear projection domains we mirror the hyperparameter selections from previous work~\cite{vassiliades:gecco18,fontaine2020covariance} and tuned manually the hyperparameters of our newly proposed algorithms: \mbox{OG-MAP-Elites}, \mbox{OG-MAP-Elites (line)}, \mbox{OMG-MEGA}, and \mbox{CMA-MEGA}. For the latent space illumination domain, since the domain is new, we manually tuned each algorithm so that the learning rate was as small as possible while still have perceptible differences in the first 10 iterations of each algorithm. We chose small learning rates because for large learning rates, the search moved far away from the training distribution, resulting in unrealistic images. We report the parameters of each algorithm below. In all algorithms we used a batch size $\lambda = 36$ following previous work~\citep{fontaine2020covariance}. MAP-Elites and the algorithms derived from MAP-Elites (MAP-Elites (line), OG-MAP-Elites, OG-MAP-Elites (line), OMG-MEGA) had an initial population size of 100, sampled from a distribution $\mathcal{N}(\bm{0},I)$. Initial solutions used to seed the initial archive do not count as an iteration in our experiments. We set the initial search position $\bm{\theta_0} = \bm{0}$ for CMA-ME, CMA-MEGA and CMA-MEGA (Adam) in all domains.

\noindent\textbf{Linear Projection (Sphere, Rastrigin).} 
\begin{itemize}
    \item MAP-Elites: $\sigma = 0.5$
    \item MAP-Elites (line): $\sigma_1 = 0.5$,  $\sigma_2 = 0.2$
    \item CMA-ME: $\sigma = 0.5$
    \item OG-MAP-Elites: $\sigma = 0.5$, $\eta = 0.5$
    \item OG-MAP-Elites (line): $\sigma_1 = 0.5$,  $\sigma_2 = 0.2$, $\eta = 0.5$
    \item OMG-MEGA: $\sigma_g = 10.0$
    \item CMA-MEGA:  $\sigma_g = 10.0$, $\eta = 1.0$ 
    \item CMA-MEGA (Adam): $\sigma_g = 10.0$, $\eta = 0.002$
\end{itemize}

\noindent\textbf{Arm Repertoire.}
\begin{itemize}
    \item MAP-Elites:  $\sigma = 0.1$
    \item MAP-Elites (line):  $\sigma_1 = 0.1$, $\sigma_2 = 0.2$
    \item CMA-ME: $\sigma = 0.2$
    \item OG-MAP-Elites: $\sigma = 0.1,\eta = 100$
    \item OG-MAP-Elites (line): $\sigma_1 = 0.1$,  $\sigma_2 = 0.2$, $\eta = 100$
    \item OMG-MEGA: $\sigma_g =1.0$
    \item CMA-MEGA: $\sigma_g= 0.05$, $\eta = 1.0$
    \item CMA-MEGA (Adam):  $\sigma_g = 0.05$, $\eta = 0.002$
\end{itemize}

\noindent\textbf{Latent Space Illumination of StyleGAN guided by CLIP.}
\begin{itemize}
    \item MAP-Elites:  $\sigma = 0.2$
    \item MAP-Elites (line):  $\sigma_1 = 0.1$, $\sigma_2 = 0.2$ 
    \item CMA-ME: $\sigma = 0.02$
    \item CMA-MEGA: $\sigma_g = 0.002$, $\eta = 1.0$ 
    \item CMA-MEGA (Adam):  $\sigma_g = 0.002$, $\eta = 0.002$ 
\end{itemize}

\noindent\textbf{Adam Hyperparameters.}
We use the same hyperparameters as the StyleGAN+CLIP implementation~\cite{styleclip}. We configure Adam with the same hyperparameters for each domain.
\begin{itemize}
    \item $\beta_1=0.9$
    \item $\beta_2=0.999$
\end{itemize}

\section{Domain Details.} \label{sec:domain_details}
\noindent\textbf{Linear Projection.} We use the linear projection domains sphere and Rastrigin from previous work~\cite{fontaine2020covariance}, and selected $n=1000$ for both domains. Each domain has a different objective function (Eq.~\ref{eq:sphere}, Eq.~\ref{eq:Rastrigin}) but identical measure functions. As in previous work~\cite{fontaine2019mapping}, we offset the objective to move the optimal location away from the center of the search space, to $x_i = 5.12 \cdot 0.4 = 2.048$.


\begin{equation} \label{eq:sphere}
sphere(x) = \sum_{i=1}^{n} {x_i} ^ {2}
\end{equation}

\begin{equation} \label{eq:Rastrigin}
Rastrigin(x) = 10 n + \sum_{i=1}^{n} [x_i ^ 2 - 10 cos(2\pi x_i)]
\end{equation}

We define the 2D measure space  with the projection of the first and second half of the components $x_i$ (see Eq.~\ref{eq:projection}). 
We bound the contribution of each component through a \textit{clip} function (Eq.~\ref{eq:clip}). which restricts the measure contribution of each $x_i$ to the range  $[-5.12, 5.12]$.


\begin{equation} \label{eq:clip}
  clip(x_i) =
  \begin{cases}
    x_i & \text{if $-5.12 \leq x_i \leq 5.12$} \\
    5.12 / x_i & \text{otherwise}
  \end{cases}
\end{equation}

\begin{equation} \label{eq:projection}
m(x) = \left( \sum_{i=1}^{\lfloor{\frac{n}{2}}\rfloor}  {clip(x_i)}, \sum_{i=\lfloor{\frac{n}{2}}\rfloor+1}^{n}  {clip(x_i)} \right)
\end{equation}

We observe that the partial derivatives for $m_i$ are $1$ for the range $[-5.12, 5.12]$. A constant derivative means that the gradient step will not shrink as OMG-MEGA approaches an extreme point in measure space, thus OMG-MEGA often overshoots the bounds; on the other hand, CMA-MEGA dynamically adapts the gradient steps, allowing for efficient coverage the measure space.  

\begin{figure}[t!]
        \centering
        \begin{tabular}{ccc}
        \begin{subfigure}{0.3\linewidth}
          \includegraphics[width = 1.0\linewidth]{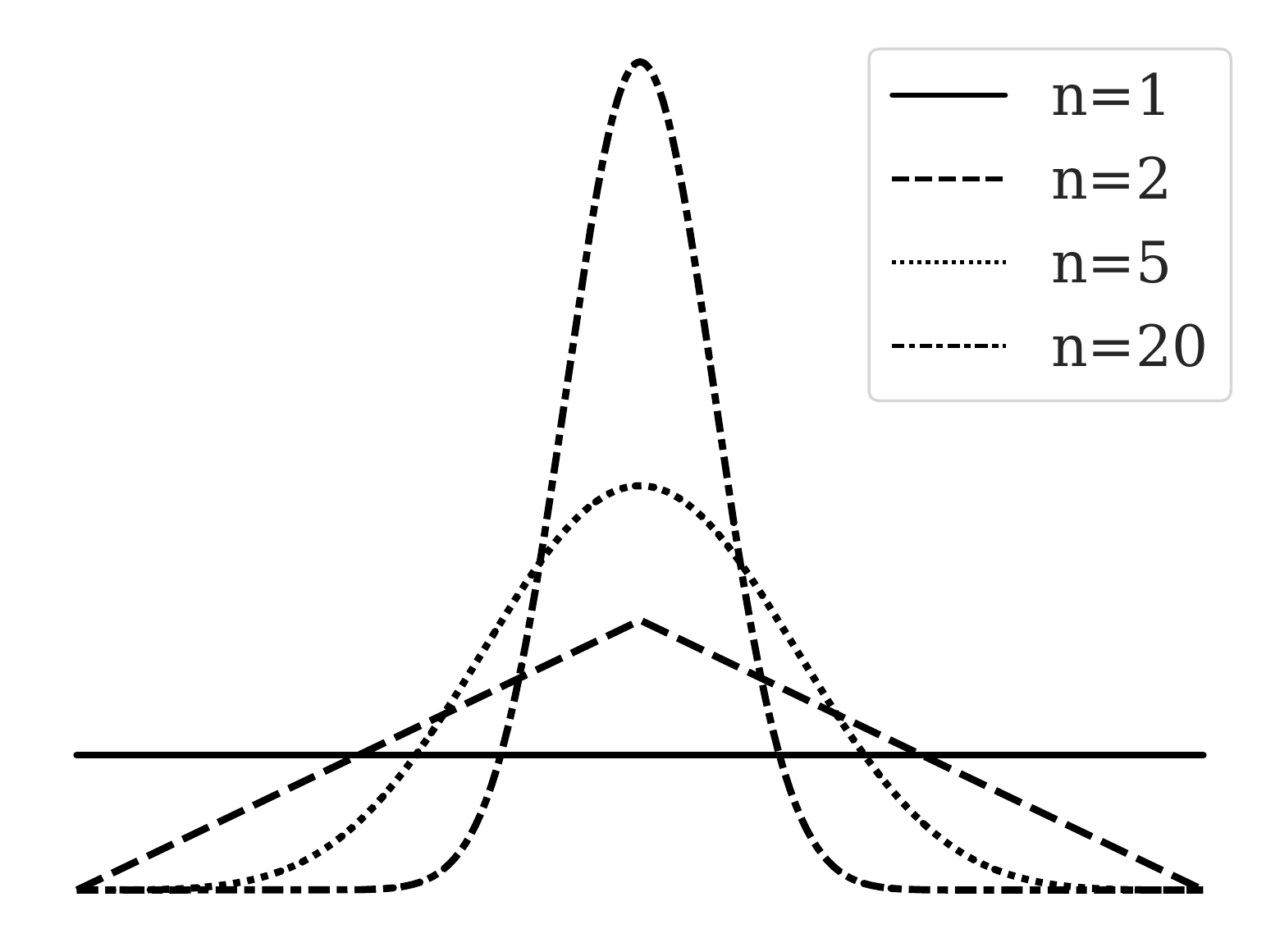}
          \caption{}
          \end{subfigure}& 
        \begin{subfigure}{0.3\linewidth}
          \includegraphics[width = 1.0\linewidth]{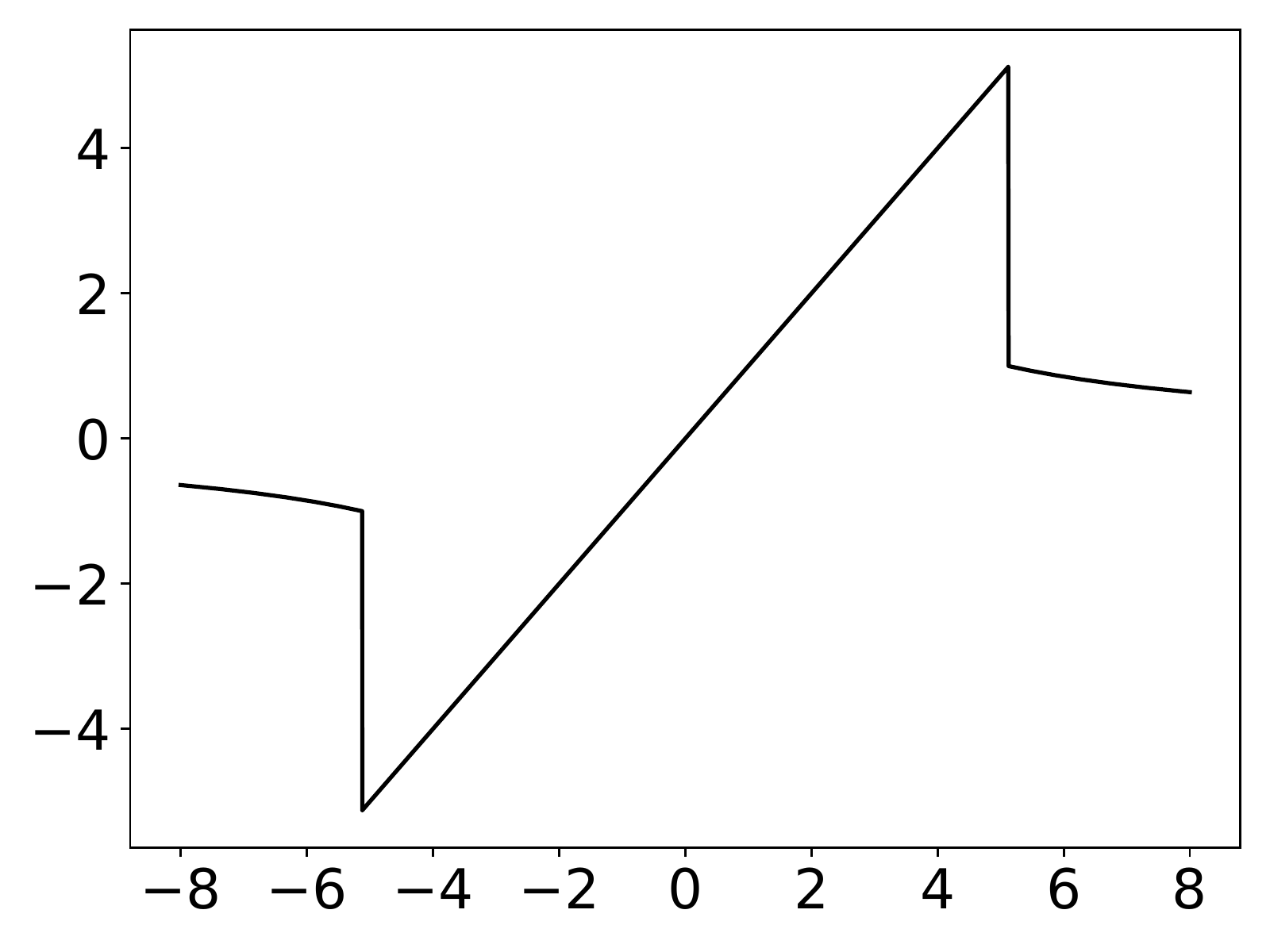}
          \caption{}
          \end{subfigure}& 
        \begin{subfigure}{0.3\linewidth}
          \includegraphics[width = 1.0\linewidth]{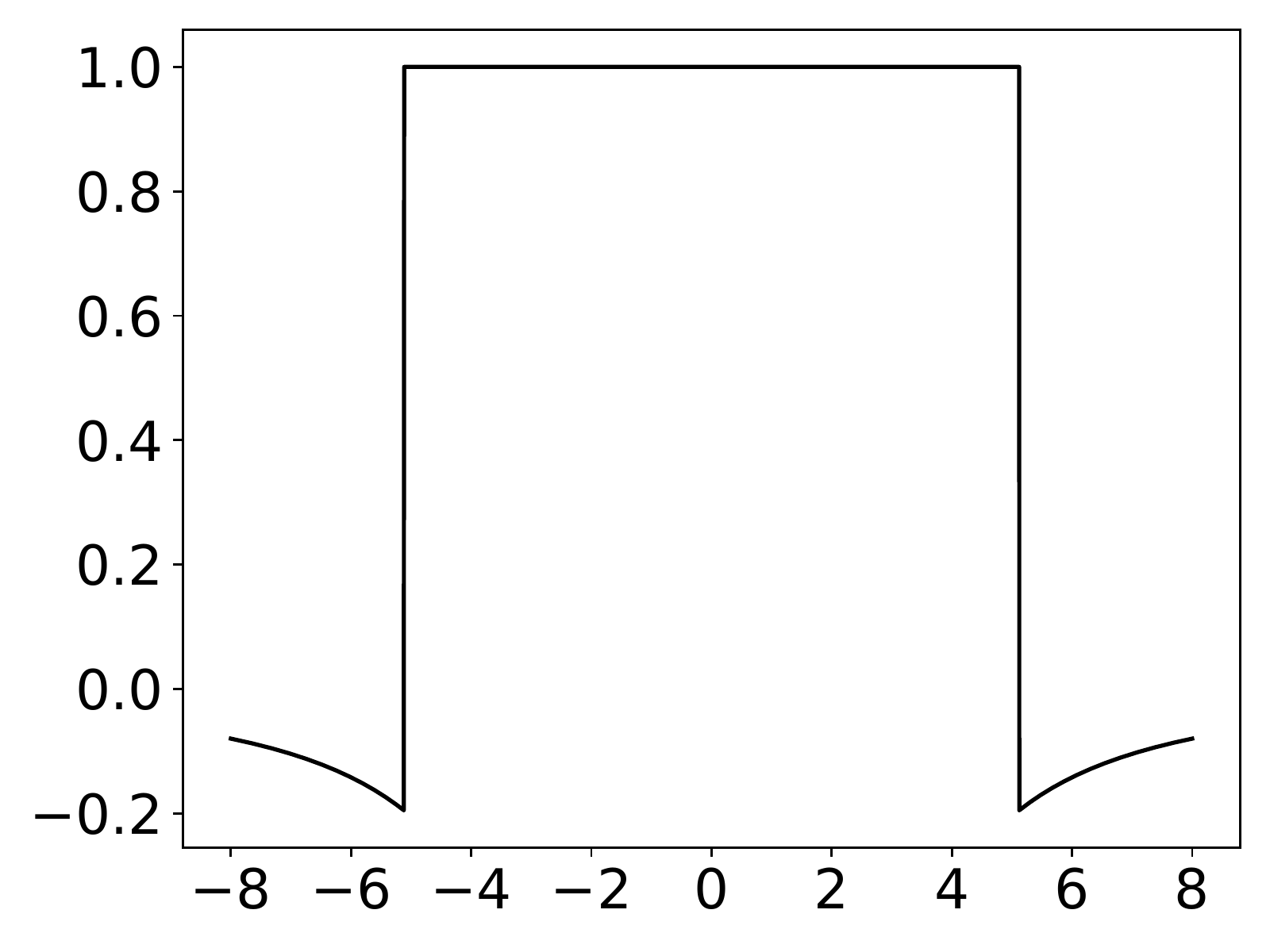}
          \caption{}
          \end{subfigure} 
         \end{tabular}
       \caption{(a) Bates distribution (reproduced from~\cite{fontaine2020covariance} with authors' permission). (b) $clip$ function for the linear projection domain. (c) Derivative of the $clip$ function.}
        \label{fig:lp_examples}
    \end{figure}

Fig.~\ref{fig:lp_examples} visualizes the challenge of the linear projection domain. Observe that if we sample uniformly on the hypercube $[-5.12, 5.12]^n$, then each of our measure functions becomes a sum of random variables. If we normalize each measure by dividing by $n$, then our measure functions become an \emph{average} of random variables. The average of $n$ random variables forms the Bates distribution~\citep{johnson1995continuous}, a distribution that narrows as $n$ increases (Fig.~\ref{fig:lp_examples}(a)). At $n=1000$ the solutions sampled from the hypercube are very close to $\bm{0}$ in measure space with high probability. A QD algorithm could simply increase its step-size to move to extremes of measure space, but the \textit{clip} function prevents this by implicitly bounding the extremes of the measure space; each component of $\bm{\theta}$ can contribute at most $\pm 5.12$ to change the position in measure space. We note the heavy penalty in the \textit{clip} function for a component leaving the range $[-5.12, 5.12]$. The combination of the narrowness of the Bates distribution and the implicit bounding of the \textit{clip} function means that a QD algorithm must dynamically \textit{adapt} its sampling distribution by shrinking step-sizes as the distribution gets close to the extremes of the measure space.

\begin{figure}[t!]
        \centering
        \begin{tabular}{ccc}
        \begin{subfigure}{0.3\linewidth}
          \includegraphics[width = 1.0\linewidth]{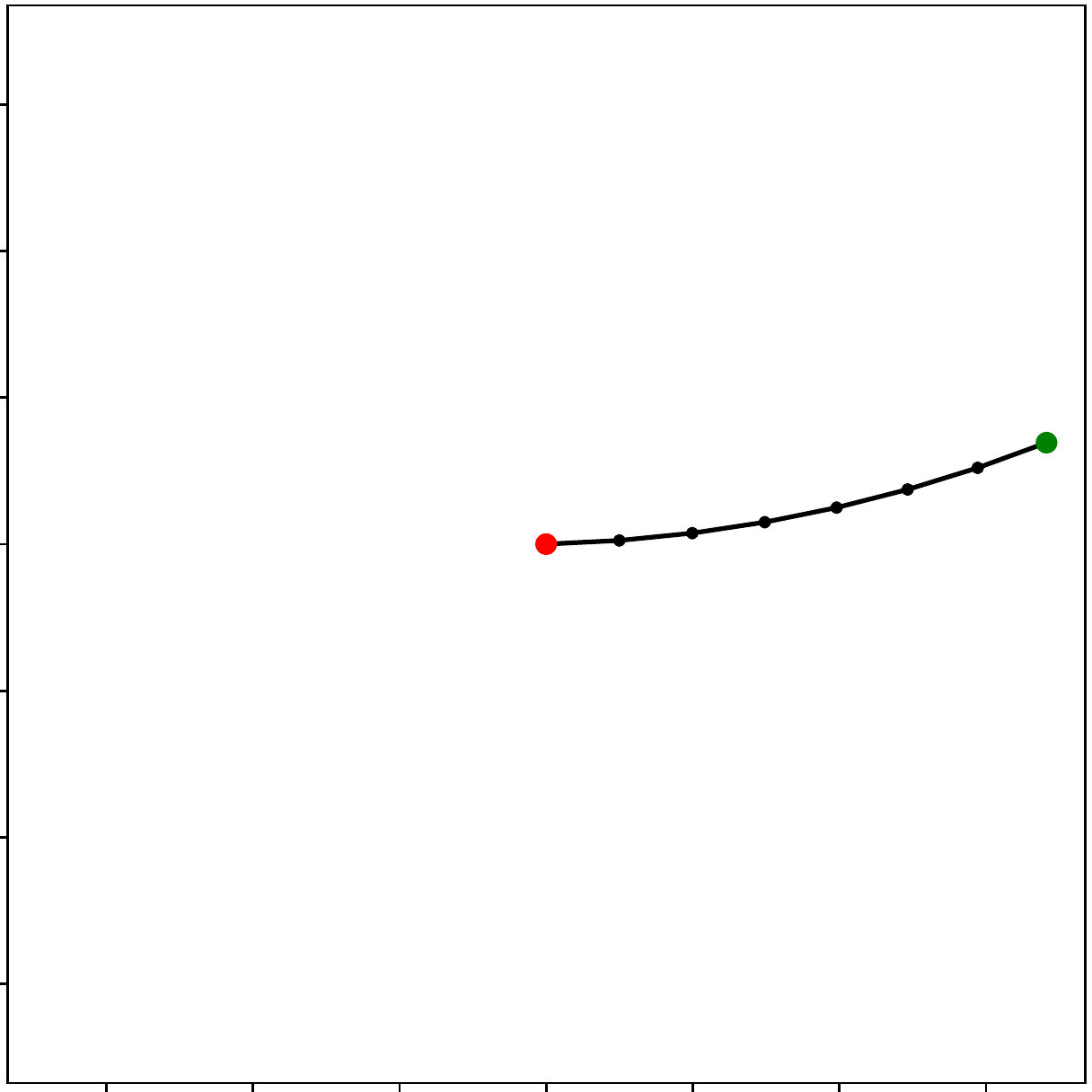}
          \caption{}
          \end{subfigure}& 
        \begin{subfigure}{0.3\linewidth}
          \includegraphics[width = 1.0\linewidth]{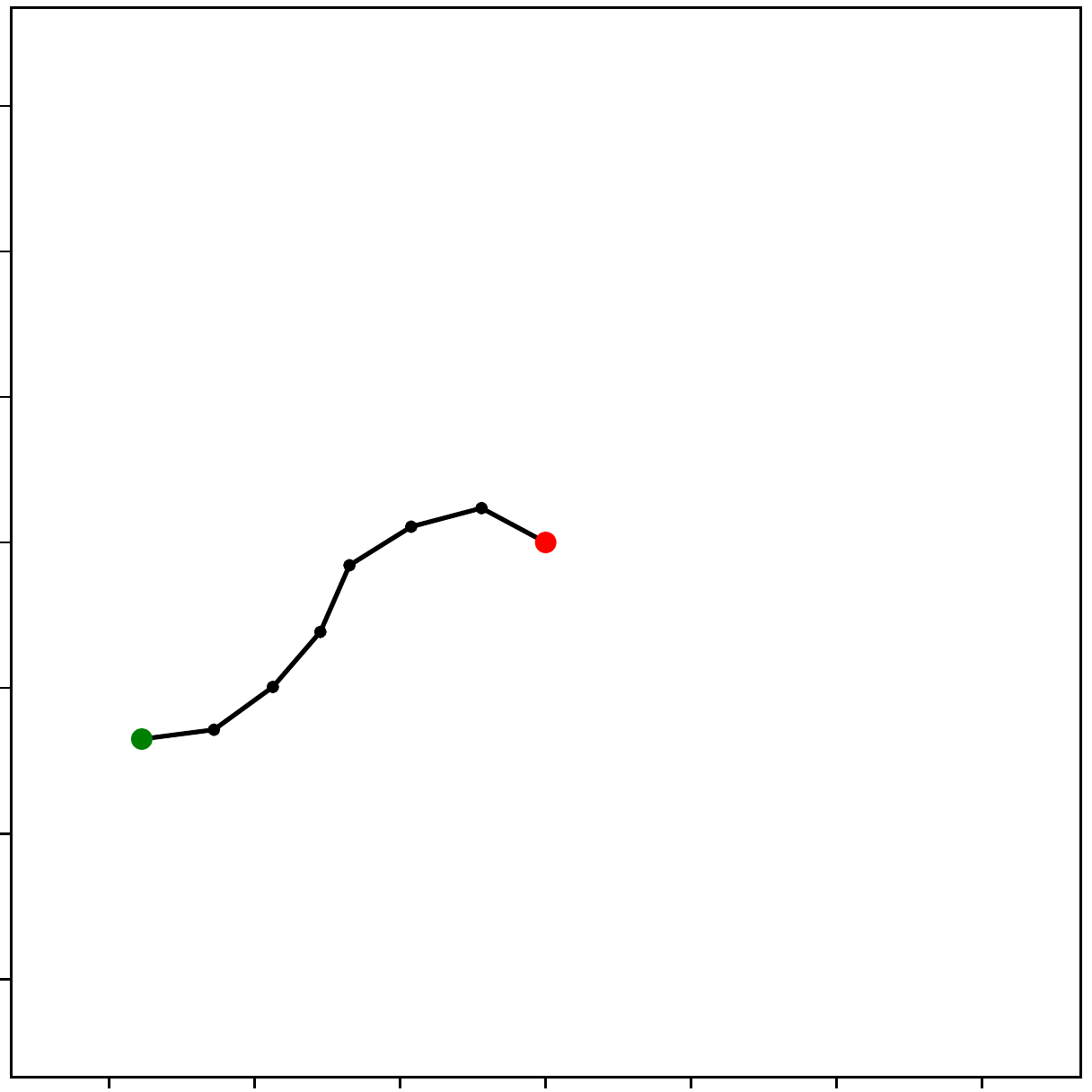}
          \caption{}
          \end{subfigure}& 
        \begin{subfigure}{0.3\linewidth}
          \includegraphics[width = 1.0\linewidth]{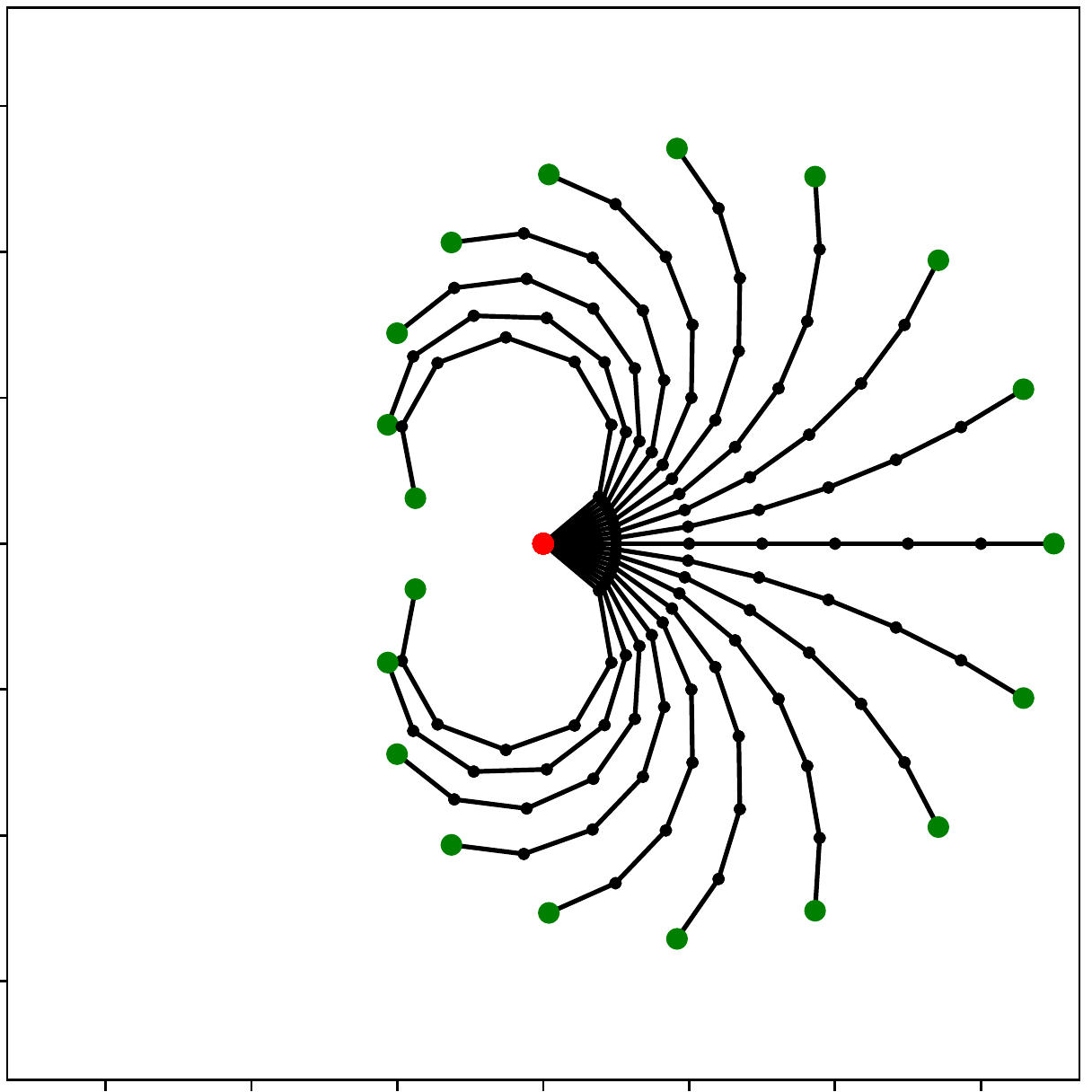}
          \caption{}
          \end{subfigure} 
         \end{tabular}
4        \caption{(a) Example of an optimal solution (zero variance of joint angles) for a 7-DOF planar arm. The green dot indicates the robot's end-effector. (b) Example of a sub-optimal solution. (c) Ensemble of optimal solutions. A precise view of the filled archive at $n=1000$ (1000-DOF), with the ``characteristic swirls'' found in lower dimensional archives, is shown in Fig.~\ref{fig:front}.}
        \label{fig:arm_examples}
    \end{figure}

\noindent\textbf{Arm Repertoire.} We visualize example solutions for a $n=7$ (7-DOF) planar arm in Fig.~\ref{fig:arm_examples}. The optimal solutions in this domain have zero joint angle variance from the mean (all angles are equal). We note that we selected as objective the variance, instead of the standard deviation used in previous work~\cite{vassiliades:gecco18}, so that the objective is differentiable at 0. The measures for the arm repertoire have range $[-n,n]$, since they are the positions of the end-effector, and the arm has $n$ links of length $l_i=1$. Thus, the reachable space forms a filled circle with radius $n$, and the maximum archive coverage becomes $\frac{\pi n^2}{4n^2} \approx 78.5\% $. We selected $n=1000$ (1000-DOF) for the experiments.


\noindent\textbf{Latent Space Illumination.} The latent space of StyleGAN~\cite{karras2019style} has size $512$ where a latent code is repeated $18$ times for each level of detail.

\noindent\textbf{Transformations of the Objective Function.} The QD-score metric, which we use to estimate the performance of QD algorithms, associates solution quality with maximizing the objective value $f$, and assumes a strictly positive objective value $f$ for every solution. Therefore, in each domain we transform the objective through a linear transformation of the form $f' = a f + b$ to map the objective values to the range $[0, 100]$.

In the linear projection domain, we follow the objective transformation proposed in previous work~\citep{fontaine2020covariance}. The original objective is to minimize the sphere and Rastrigin functions. We compute an estimate on the largest sphere and Rastrigin values in the hypercube $[-5.12, 5.12]^n$ which contain only components in the linear portion of the clip function. We compute $f(-5.12, -5.12, ..., -5.12)$ as an estimated maximum of the function $f_{max}$. The minimum of each function is $f_{min}=0ß$. We then remap the function values for both the sphere and Rastrigin objective functions through the linear transformation given by:

\begin{equation}
f'(\bm{\theta}) = 100 \cdot \frac{f(\bm{\theta}) - f_{max}}{f_{min} - f_{max}}
\label{eq:obj_transform}
\end{equation}

For the arm domain we estimate $f_{max}$ to be $1$ from an initial population of angles sampled from $\mathcal{N}(\bm{0},I)$. For the LSI domain we picked $f_{max} = 10$ by observing the CLIP loss function values for the objective text prompt ``Elon Musk with short hair.''.

\section{Implementation}
\noindent\textbf{Archives.} In the both the linear projection and arm repertoire domains, the measure space is 2D, with resolution $100 \times 100$. We normalize the objective value $f$ so that it is in the range [0,100] and initialize empty cells of the archive with an objective value $f$ of 0. For latent space illumination we form a 2D archive with the resolution $200 \times 200$. We double the resolution in each dimension as the examined QD algorithms fill roughly a quarter of possible cells in the archive.

\noindent\textbf{Computational Resources.} We ran 10,000 iterations in all algorithms. We ran all trials in parallel on an AMD Ryzen Threadripper 32-core (64 threads) processor and an GeForce RTX 3090 Nvidia GPU. A run of 20 trials in parallel required about 30' for the linear projection domains and 2 hours for the arm repertoire domain. One trial run for the latent space illumination domain took about 2 hours. We note runtime increases at a higher logging frequency and algorithms which perform better may run slower due to iterating over more archive solutions when QD statistics are calculated.

\noindent\textbf{Software Implementation.} We use the publicly available Pyribs~\citep{pyribs} library for all algorithms, where we implemented the OG-MAP-Elites, OG-MAP-Elites (line), OMG-MEGA and CMA-MEGA algorithms. We use the Adam implementation of ESTool.~\footnote{\url{https://github.com/hardmaru/estool/blob/master/es.py} (line 70)}

\section{Improvement Ranking as a Natural Gradient Approximation}

We show for the first time that CMA-ME's improvement ranking optimizes a modified QD objective (eq.~\ref{eq:mod_objective}) via a natural gradient approximation. We then extend our derivation to show that the coefficient distribution of CMA-MEGA approximates natural gradient steps of the same objective with respect to the gradient coefficients.


We let the original QD objective for an archive $\mathcal{A}$ be: 
\begin{equation}
\max \ J_1(\mathcal{A}) =  \sum_{i=1}^M f(\bm{\theta_i})
\label{eq:orig_objective}
\vspace{-1em}
\end{equation}

where $f(\bm{\theta_i}) = 0$ if a cell $i$ is unfilled.



Consider that MAP-Elites solves a discrete optimization problem over \textit{archives} $\mathcal{A}$ rather than a continuous optimization problem over solution parameters $\bm{\theta}$. To analyze the improvement ranking, we reparameterize the objective $J_1$ to include $\bm{\theta}$, the current search position for CMA-ES. By fixing the current archive $\mathcal{A}$, our goal is to show that improvement ranking approximates a natural gradient step of $J_1$ with respect to $\bm{\theta}$. In other words, we want to calculate the direction to change $\bm{\theta}$ that results in the steepest increase of $J_1$ after $\bm{\theta}$ is added to the archive.




Ranking candidate solutions $\bm{\theta}$ based on $J_1$ may prioritize solutions that improve existing cells on the archive over solutions that discover new cells. To show this possibility, let $\bm{\theta_i}$ and $\bm{\theta_j}$ be two arbitrary candidate solutions, where $\bm{\theta_i}$  replaces a cell with an existing occupant  $\bm{\theta_p}$ and $\bm{\theta_j}$ discovers a new cell. If $f(\bm{\theta_i})-f(\bm{\theta_p}) > f(\bm{\theta_j})$, then $\bm{\theta_i}$ will be ranked higher than $\bm{\theta_j}$.

To strictly prioritize exploration, CMA-ME performs a two-stage improvement ranking on the objective $J_1$, where it ranks first all solutions  $\bm{\theta_j}$ that discover new cells based on their objective value $f( \bm{\theta_j})$, and subsequently all solutions $\bm{\theta_i}$ that improve existing cells based on the difference  $f(\bm{\theta_i})-f(\bm{\theta_p})$ between the $f$ value of the new and previous solution $\bm{\theta_p}$.

To show that the ranking rules of CMA-ME are equivalent to a natural gradient of a QD objective, we specify a function $J_2$ (see Eq.~\ref{eq:mod_objective}), and we will show that \emph{sorting solutions purely by $J_2$ results in the same order as the improvement ranking by CMA-ME on the objective $J_1$.}



\begin{equation}
\max \ J_2(\mathcal{A}) =  \sum_{i=1}^M [f(\bm{\theta_i})+b_i C]
\label{eq:mod_objective}
\end{equation}

\begin{equation}
C > 2 \cdot [\max_{\bm{\theta_i} \in \mathbb{R}^n} f(\bm{\theta_i}) -\min_{\bm{\theta_i} \in \mathbb{R}^n} f(\bm{\theta_i})]
\label{eq:define_c}
\end{equation}

where $C$ is larger than two times the largest gap between any pair of $f(\bm{\theta}_i)$ and $f(\bm{\theta}_j)$ (see Eq.~\ref{eq:define_c}), $b_i$ is 1 if a cell is occupied and 0 otherwise, and $f(\bm{\theta_i}) = 0$ if a cell $i$ is unfilled.\footnote{We note that this is an initialization value, since an empty cell does not yet contain a $\bm{\theta_i}$.}





Let $\bm{\theta_i}$ and $\bm{\theta_j}$ be two arbitrary candidate solutions whose addition to the archive changes the archive. 
We define  $\Delta^{\mathrm{J_2}}_i = J_2(\mathcal{A}+\bm{\theta_i}) - J_2(\mathcal{A}) $ 
and $\Delta^{\mathrm{J_2}}_j = J_2(\mathcal{A}+\bm{\theta_j}) - J_2(\mathcal{A}) $, and we consider the following three cases:



\noindent\textbf{1: Without loss of generality $\bm{\theta_i}$ improves an occupied cell and $\bm{\theta_j}$ discovers a new cell.}

Based on the improvement ranking of CMA-ME, $\bm{\theta_j}$ will always be ranked higher than $\bm{\theta_i}$. We will show that the same ordering holds if we sort based on the objective $J_2$.

We let $\bm{\theta_p}$ be the occupant replaced by $\bm{\theta_i}$.

We show that $\Delta^{\mathrm{J_2}}_j >\Delta^{\mathrm{J_2}}_i$:

\begin{align}
\begin{split}
\Delta^{\mathrm{J_2}}_j &= f(\bm{\theta_j}) + C \\ &=
f(\bm{\theta_j}) + \frac{C}{2} +\frac{C}{2}  \\ &>  f(\bm{\theta_j}) + \frac{C}{2} + f(\bm{\theta_i})-f(\bm{\theta_p}) \\ &> f(\bm{\theta_i})-f(\bm{\theta_p}) \\ & = \Delta^{\mathrm{J_2}}_i
\end{split}
\end{align}

Where inequalities hold from the definition of $C$ (Eq.~\ref{eq:define_c}).

\noindent\textbf{2: Both $\bm{\theta_i}$ and $\bm{\theta_j}$ discover new cells.}

From the improvement ranking rule, $\bm{\theta_i}$ will be ranked higher than $\bm{\theta_j}$ iff $\Delta^{\mathrm{J_1}}_i>\Delta^{\mathrm{J_1}}_j$.

We show that this is equivalent to $\Delta^{\mathrm{J_2}}_i>\Delta^{\mathrm{J_2}}_j$.

\begin{align}
\begin{split}
\Delta^{\mathrm{J_1}}_i > \Delta^{\mathrm{J_1}}_j \Leftrightarrow 
f(\bm{\theta_i}) >  f(\bm{\theta_j}) \Leftrightarrow 
f(\bm{\theta_i})+C >  f(\bm{\theta_j}) +C  \Leftrightarrow 
\Delta^{\mathrm{J_2}}_i > \Delta^{\mathrm{J_2}}_j 
\end{split}
\end{align}

\noindent\textbf{3: Both $\bm{\theta_i}$ and $\bm{\theta_j}$ improve existing cells.}

 
We let  $\bm{\theta_p}$ the occupant replaced by $\bm{\theta_i}$ and  $\bm{\theta_r}$ the occupant replaced by $\bm{\theta_j}$.

\begin{align}
\begin{split}
\Delta^{\mathrm{J_1}}_i > \Delta^{\mathrm{J_1}}_j \Leftrightarrow  f(\bm{\theta_i}) - f(\bm{\theta_p}) >  f(\bm{\theta_j})   - f(\bm{\theta_r}) \Leftrightarrow 
\Delta^{\mathrm{J_2}}_i > \Delta^{\mathrm{J_2}}_j 
\end{split}
\end{align}

Therefore, the CMA-ME improvement ranking is identical to ranking purely based on $J_2$. 

Previous work~\cite{akimoto2010bidirectional} has shown that CMA-ES approximates the natural gradient of its provided optimization objective function to a scale. At the same time, CMA-ES is invariant to order-preserving transformations of its objective function, since it is a comparison-based optimization method~\cite{hansen:cma16}. We provide an explicit objective function $J_2$, and we have shown that maximizing this function results in the same ordering as the one specified implicitly by the improvement ranking rules. Therefore, CMA-ME, which uses improvement ranking to update the CMA-ES parameters, approximates a natural gradient of $J_2$ to a scale.

CMA-MEGA sorts solutions via the same improvement ranking rules as CMA-ME to update an internal sampling distribution maintained by CMA-ES. Unlike CMA-ME, CMA-MEGA updates a sampling distribution of gradient step coefficients, and not solution parameters directly. Therefore, CMA-MEGA performs a form of adaptive search, where the algorithm dynamically changes the search hyperparameters (gradient step coefficients) to maximize $J_2$.

By connecting CMA-ME to previous work that connects natural gradient descent and CMA-ES~\cite{akimoto2010bidirectional}, we gain a new perspective on both CMA-ME and CMA-MEGA. In each iteration, CMA-ME optimizes for a single solution whose addition to the archive results in the largest increase in $J_2$. However, after sampling, we update the archive which results in a small change to the optimization landscape. In other words, CMA-ME assumes that CMA-ES is a robust enough optimizer to handle small changes to its target objective.

To apply this new perspective to CMA-MEGA, we observe that CMA-MEGA solves the optimization problem of finding the chosen gradient coefficients that yield the largest increase in $J_2$ for a fixed $\bm{\theta}$ and archive $\mathcal{A}$. In other words, CMA-MEGA optimizes the objective $J_2$, as CMA-ME, but it does so directly in objective-measure space. However, after taking one optimization step, we change the single solution optimization problem by updating the archive and moving $\bm{\theta}$. Just like CMA-ME, we assume that CMA-ES is a robust enough optimizer to handle changes in the optimization landscape brought about by changing the archive and $\bm{\theta}$. 

\begin{figure}[t!]
\centering
\includegraphics[width=0.8\linewidth]{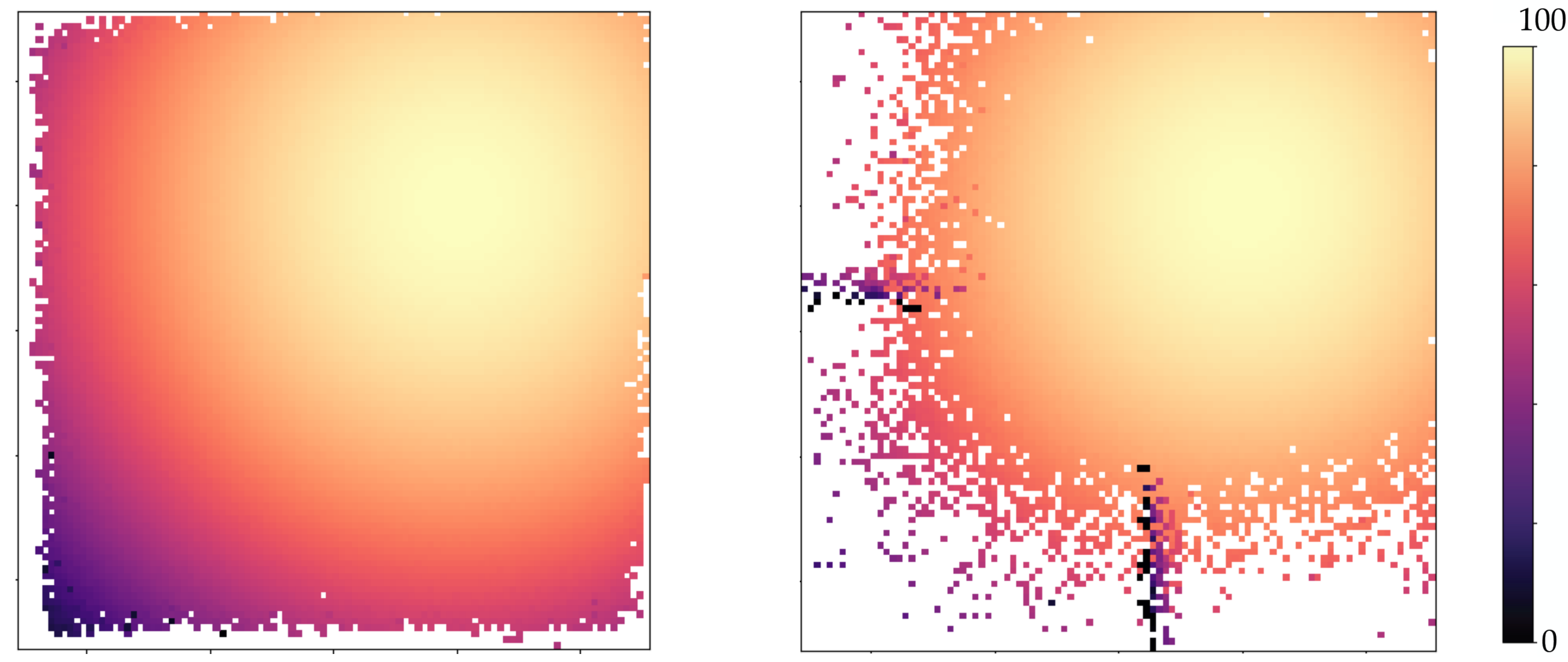}
\caption{Example archive of OMG-MEGA for normalized (left) and unnormalized (right) gradients for the ablation run in the Linear Projection (sphere) domain.}
\label{fig:ablation_norm_gradients}
\end{figure}

\section{On the Importance of Normalizing Gradients} 
\begin{table}
\centering
\resizebox{0.7\linewidth}{!}{
\begin{tabular}{l|rrr}
             & \multicolumn{3}{c}{LP (sphere)}   \ \\ 
    \toprule
Algorithm     & QD-score & Coverage  & Best \\
    \midrule
OMG-MEGA (norm)  & 71.58 $\pm$ 0.10 & 92.09 $\pm$ 0.21\% & 100.00 $\pm$ 0.00 \\
OMG-MEGA (unnorm)  &  56.96 $\pm$ 0.24 & 67.30 $\pm$ 0.38\%  &  100.00 $\pm$ 0.00\\
\bottomrule
\end{tabular}
}
\resizebox{0.7\linewidth}{!}{
\begin{tabular}{l|rrr}
             & \multicolumn{3}{c}{LP (Rastrigin)}   \ \\ 
    \toprule
Algorithm     & QD-score & Coverage  & Best \\
    \midrule
OMG-MEGA (norm)   & 55.90 $\pm$ 0.21 & 77.00 $\pm$ 0.34\% & 97.55 $\pm$ 0.06 \\
OMG-MEGA (unnorm)  & 16.56  $\pm$ 0.11& 25.97  $\pm$ 0.18\%  &  89.76$\pm$ 0.42 \\
\bottomrule
\end{tabular}
}
\resizebox{0.7\linewidth}{!}{
\begin{tabular}{l|rrr}
             & \multicolumn{3}{c}{Arm Repertoire}   \ \\ 
    \toprule
Algorithm     & QD-score & Coverage  & Best \\
    \midrule
OMG-MEGA (norm) & 44.12 $\pm$ 0.06 & 44.13 $\pm$ 0.06\% & 100.00 $\pm$ 0.00 \\
OMG-MEGA (unnorm)  &  0.85 $\pm$ 0.04 & 7.02 $\pm$ 0.21\%  & 12.58 $\pm$ 0.36\\
\bottomrule
\end{tabular}
}\vspace{1em}
\caption{OMG-MEGA with normalized (norm) and unnormalized (unnorm) gradients.} 
\label{tab:unnorm_comparison}
\end{table}

We discuss the importance of normalizing gradients in the MAP-Elites via Gradient Arborescence algorithm and include results from an ablation over gradient normalization. We explore the importance of normalizing gradients through an ablation study on OMG-MEGA, where we compare \mbox{OMG-MEGA} with the normalization step to OMG-MEGA without the normalization step in the linear projection and arm repertoire domains.

To account for the scaled step-size after normalization in the linear projection domain, we set the $\sigma_g$ to $10$ in the normalized variant and to $0.5$ in the unnormalized variant, since the average gradient magnitude was approximately $20$. For the arm repertoire domain, we note that the objective gradient magnitude was approximately $10^{-2}$, while the measure gradient magnitudes were approximately $10^2$. We ran tuning experiments for $\sigma_g$ at scales $10^2$, $10^1$, $1$, $10^{-1}$, $10^{-2}$, $10^{-3}$, and $10^{-4}$ to scale the unnormalized gradient and selected $\sigma_g = 10^{-4}$ as the best performing hyperparameter. 

Table~\ref{tab:unnorm_comparison} presents the results of the ablation study in the linear projection and arm domains. The quantitative results suggest that normalization greatly affects performance of the OMG-MEGA algorithm. Below we give our interpretation of why normalization plays such a significant factor on performance in each domain.

Consider the linear projection sphere domain. In this domain the objective is quadratic. As we move further away from the optimum, the magnitude of the gradient increases linearly. In OMG-MEGA, the gradient coefficients are sampled from a fixed distribution. This means the gradient of our objective will dominate the gradients of the measures for positions in measure space far away from the optimum. Fig.~\ref{fig:ablation_norm_gradients} visualizes a final archive from this domain. We see that the unnormalized OMG-MEGA easily fills cells near the global optimum, but struggles to fill cells with a lower objective value. In contrast OMG-MEGA with normalization fills the archive evenly.


In the arm repertoire domain, we observe that the objective gradient is on a different scale than the measure gradients and this affects the performance of the unnormalized OMG-MEGA. The average objective gradient magnitude was $10^{-2}$. In contrast, the average measure gradient magnitude was $10^2$. This means that  the objective gradient barely contributes to the search for a fixed coefficient distribution. However, the fact that OG-MAP-Elites performs well in this domain indicates that the objective gradient is an important factor in filling the archive. Because of the different scale, the objective gradient is largely ignored by the OMG-MEGA variant that does not normalize gradients.


While in OMG-MEGA the benefit of normalizing gradients is clear, we speculate that normalizing gradients is beneficial also for CMA-MEGA. In CMA-MEGA, the CMA-ES subroutine solves a non-stationary optimization problem for gradient coefficients that maximize the QD objective. The coefficients can be viewed as a learning rate for each gradient. In this sense, CMA-MEGA has adaptive learning rates, where CMA-ES acts as the adaptation mechanism. However, in standard gradient ascent the magnitude of the gradient acts as a step size control. If the gradients are not normalized, the magnitudes and the adaptive step size control may end up fighting each other. This may cause oscillations in CMA-ES's adaptation mechanisms and lead to instability. We leave a thorough study to evaluate the exact benefits of normalization in CMA-MEGA as future work. 

Finally, we note a conceptual difference for how we leverage measure gradients in DQD versus how objective gradients are used in optimization. In quality diversity, a goal of the algorithm is to cover the entire measure space. Intermediate solutions are of equal importance to solutions at the extremes of measure space. By normalizing gradients and giving control over the step-size to the DQD algorithm, CMA-MEGA can adapt its search distribution to more easily fill intermediate regions of the measure space, rather than \textit{only} optimize for extrema.



\section{On the Differences between OG-MAP-Elites (line) and PGA-MAP-Elites}

We explore the design decisions between \mbox{PGA-MAP-Elites}~\citep{nilsson2021policy}, an reinforcement learning (RL) algorithm, and OG-MAP-Elites (line), a DQD baseline that draws insights from PGA-MAP-Elites. 

Specifically, PGA-MAP-Elites is an actor-critic reinforcement learning algorithm. Actor-critic methods combine elements of policy gradient methods, which approximate the gradient of the reward function with respect to a policy's actions, and Q-learning methods, which approximate \mbox{Q-values} for every state-action pair. The computationally expensive part in this setting is the evaluation of a policy, while gradient computations do not factor into the running time of the algorithm. 

On the other hand, the DQD problem assumes knowledge of the exact gradients of the objective and measure functions, which makes gradient approximation steps unnecessary. Thus, the most expensive part of the computation in many applications becomes the gradient computation for a given solution, rather than the evaluation of that solution. For example, if the function is a neural network, evaluation of a solution can be done efficiently with a forward pass. However, gradient computation requires an often expensive backpropagation of the loss function through the network. For efficient performance, a designer of a DQD algorithm would aim towards minimizing the number of gradient computations to achieve good results, instead of \textit{only} the number of evaluations. 

As an actor-critic method, PGA-MAP-Elites trains a Q-value approximator (a critic). As a quality diversity algorithm, the critic network can be trained from the rollout data of the entire \textit{archive} of policies to form a better Q-value approximation. Since an accurate gradient approximation is important in RL settings,  the shared critic network allows for efficient and accurate policy gradient calculation \textit{on demand} for any new candidate policy evaluated. PGA-MAP-Elites consists of two independent operators, (1) the Iso+LineDD operator and (2) a policy gradient step.

In our implementation of the DQD algorithms in Pyribs' ask-tell interface -- adopted from Pycma~\cite{hansen2019pycma}, we evaluate a solution and pass the gradient simultaneously. If we directly implemented independent operators in OG-MAP-Elites (line), when using operator (2) we would need to evaluate a solution once to perform a gradient step and evaluate the new solution again after the gradient step. Instead, we apply in OG-MAP-Elites (line) sequentially the operators (1) and (2). This allows us to use the existing evaluation (and gradient computation) from (1) to perform a gradient step (2), reducing the number of evaluations needed per gradient step.

We implemented OG-MAP-Elites and OG-MAP-Elites (line) with the two independent operators to evaluate the effect of this design decision. We can divide our computation budget per iteration into thirds. The first third of operators will perturb existing solutions with Iso+LineDD. The second third will re-evaluate an existing archive solution and compute an objective gradient. The final third will evaluate the solution after applying the computed gradient step. With our current derivation, we waste a third of our evaluation budget on computing gradients. We denote implementations with independent operators as OG-MAP-Elites$^\dag$ and OG-MAP-Elites (line)$^\dag$. 
 

We run our ablation on the arm repertoire domain and both variants of the linear projection domain. Table~\ref{tab:ogmeline_independent} contains the results of the ablation. On both linear projection domains, the variants of \mbox{OG-MAP-Elites} and \mbox{OG-MAP-Elites (line)} with sequential operators outperform their counterparts with independent operators. However, in the arm repertoire domain, the independent operators implementation outperforms the sequential operator counterpart. We conjecture that because exploration is more difficult in the linear projection domain, the smaller number of perturbation operations in the independent operators variants hurts performance. On the other hand,  in the arm repertoire domain, optimal solutions are concentrated in a small elite hypervolume. The independent operators can apply two or more gradient steps sequentially, directing the search efficiently towards the elite hypervolume.


\begin{table}
\centering

\resizebox{0.7\linewidth}{!}{
\begin{tabular}{l|rrr}
             & \multicolumn{3}{c}{LP (sphere)}   \ \\ 
    \toprule
Algorithm     & QD-score & Coverage  & Best \\
    \midrule
OG-MAP-Elites & 1.36 $\pm$ 0.08 &  1.50 $\pm$ 0.09\% & 100.00 $\pm$ 0.00 \\
OG-MAP-Elites (line)  &  14.91$\pm$ 0.08 &  17.29 $\pm$ 0.09\%  &  100.00 $\pm$ 0.00\\
OG-MAP-Elites$^\dag$ & 1.15 $\pm$ 0.10 & 1.27 $\pm$ 0.11\% & 100.00 $\pm$ 0.00 \\
OG-MAP-Elites (line)$^\dag$ & 13.31 $\pm$ 0.09 & 15.42  $\pm$ 0.10\% & 100.00 $\pm$ 0.00 \\
\bottomrule
\end{tabular}
}

\resizebox{0.7\linewidth}{!}{
\begin{tabular}{l|rrr}
             & \multicolumn{3}{c}{LP (Rastrigin)}   \ \\ 
    \toprule
Algorithm     & QD-score & Coverage  & Best \\
    \midrule
OG-MAP-Elites & 0.83 $\pm$ 0.03 & 1.28  $\pm$ 0.04\% & 74.38 $\pm$ 0.04 \\
OG-MAP-Elites (line)  &  6.09 $\pm$ 0.04 &  8.84 $\pm$ 0.07\%  &  76.67 $\pm$ 0.05\\
OG-MAP-Elites$^\dag$ & 0.76 $\pm$ 0.03 & 1.14 $\pm$ 0.03\% & 74.43 $\pm$ 0.04 \\
OG-MAP-Elites (line)$^\dag$ & 5.14 $\pm$ 0.04 & 7.46 $\pm$ 0.06\% & 76.17 $\pm$ 0.04 \\
\bottomrule
\end{tabular}
}

\resizebox{0.7\linewidth}{!}{
\begin{tabular}{l|rrr}
             & \multicolumn{3}{c}{Arm Repertoire}   \ \\ 
    \toprule
Algorithm     & QD-score & Coverage  & Best \\
    \midrule
OG-MAP-Elites & 57.21 $\pm$ 0.03  &  58.11 $\pm$ 0.03\% & 98.63 $\pm$ 0.00 \\
OG-MAP-Elites (line)  & 59.57 $\pm$  0.03 & 60.19 $\pm$ 0.03\%  &  99.17 $\pm$ 0.00\\
OG-MAP-Elites$^\dag$ & 65.72 $\pm$ 0.06 & 65.95 $\pm$ 0.06\% & 100.00 $\pm$ 0.00 \\
OG-MAP-Elites (line)$^\dag$ & 65.73 $\pm$ 0.04 & 65.96 $\pm$ 0.04\% & 100.00 $\pm$ 0.00 \\
\bottomrule
\end{tabular}
}

\vspace{1em}
\caption{OG-MAP-Elites with independent gradient and perturbation operators (denoted with $^\dag$) compared against our sequential operators derivation from the main paper.} 
\label{tab:ogmeline_independent}
\end{table}

\section{On the Effect of the Archive Resolution on the Performance of CMA-ME}

In this section, we comment on the performance of \mbox{CMA-ME} on the linear projection domain in the main paper and include an extra experiment with a different archive resolution.

The linear projection domain was first introduced by the authors of CMA-ME~\citep{fontaine2020covariance} to show the importance of adaptation mechanisms in quality diversity algorithms. In the paper, each domain was run with search dimensions of $n=20$ and $n=100$ for both the sphere and the Rastrigin objective. Experiments were run for $2.5 \times 10^6$ evaluations ($\approx 65,000$ iterations) and, due to the large number of evaluations, the archive resolution was $500 \times 500$. The paper reports \mbox{CMA-ME} outperforming both \mbox{MAP-Elites} and \mbox{MAP-Elites (line)} in this setting. In contrast, our experiments were run for 10,000 iterations with a search space dimension of $n=1000$ and an archive resolution of $100 \times 100$.

\begin{figure*}[t!]
\centering
\includegraphics[width=0.7\columnwidth]{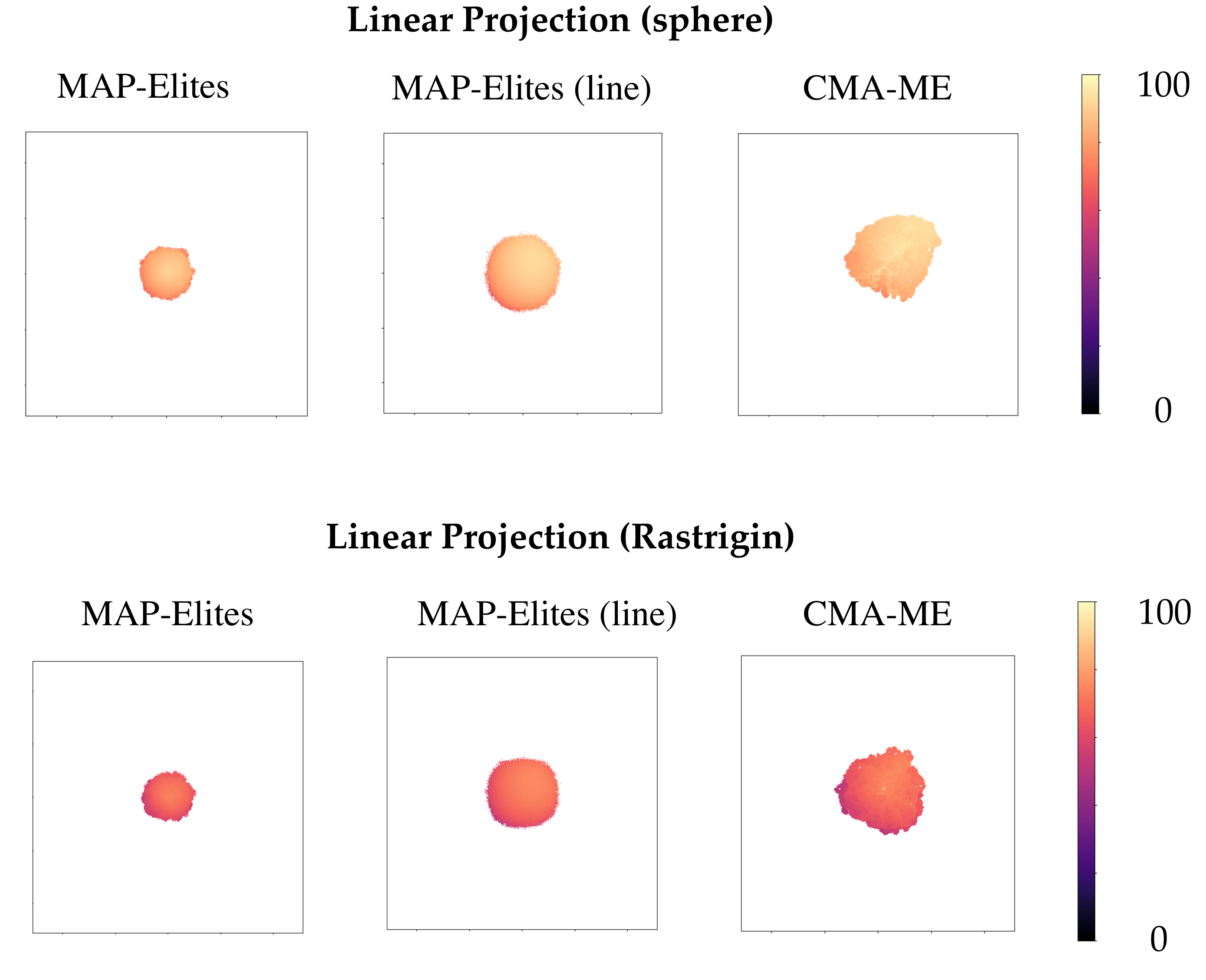}
\caption{Example archives for each derivative-free QD algorithm evaluated in the linear projection domains. These experiments were run with archives of resolution $500 \times 500$ instead of $100 \times 100$.}
\label{fig:cmame_extra_results}
\end{figure*}

We ran two additional experiments where we evaluated \mbox{MAP-Elites}, \mbox{MAP-Elites (line)}, and \mbox{CMA-ME} on both linear projection domains with archive resolution of $500 \times 500$. Table~\ref{tab:cmame_extra} contains the results of the extra experiments and Fig.~\ref{fig:cmame_extra_results} visualizes a final archive of each algorithm for each experiment. We observe that under increased archive resolution CMA-ME outperforms both variants of MAP-Elites. We also observe that coverage increases for both MAP-Elites and CMA-ME when the archive is higher resolution. In contrast, coverage decreases for the MAP-Elites (line) algorithm when the archive resolution increases. 



We explain why the low resolution of the archive negatively affects the performance of CMA-ME. For archives of resolution $100 \times 100$ there is not enough granularity in the archive to easily move through measure space, negatively affecting both MAP-Elites and CMA-ME. Moreover, CMA-ME restarts frequently due to the rule that the algorithm will restart from a random elite if none of the $\lambda$ sampled solutions in an iteration improve the archive. Everytime CMA-ME restarts, it samples $\lambda$ solutions by perturbing an elite chosen uniformly at random, with isotropic Gaussian noise. This limit case is nearly equivalent to the MAP-Elites algorithm, with the caveat that all generated solution come from one elite rather than $\lambda$ different elites. In the $100 \times 100$ setting, CMA-ME performs similarly to MAP-Elites.


However, at a higher archive resolution of $500 \times 500$ restarts become less frequent and generated solutions in each iteration fall into different cells in the archive. The higher granularity of the archive causes CMA-ME to restart fewer times. Under this setting we see performance of CMA-ME consistent with the CMA-ME paper. 

Overall, this study indicates the effect of archive resolution on the performance of QD algorithms.

\begin{table}
\centering
\resizebox{0.7\linewidth}{!}{
\begin{tabular}{l|rrr}
             & \multicolumn{3}{c}{LP (sphere)}   \ \\ 
    \toprule
Algorithm     & QD-score & Coverage  & Best \\
    \midrule
MAP-Elites &  2.37 $\pm$ 0.01 & 2.86 $\pm$ 0.01\%  &  90.56 $\pm$ 0.02\\
MAP-Elites (line) & 4.96 $\pm$ 0.03 & 5.82  $\pm$ 0.04\% &  92.87 $\pm$ 0.03\\
CMA-ME & 6.82 $\pm$ 0.05 & 7.74 $\pm$ 0.06\% & 94.80 $\pm$ 0.07\\
\bottomrule
\end{tabular}
}

\resizebox{0.7\linewidth}{!}{
\begin{tabular}{l|rrr}
             & \multicolumn{3}{c}{LP (Rastrigin)}   \ \\ 
    \toprule
Algorithm     & QD-score & Coverage  & Best \\
    \midrule
MAP-Elites & 1.87 $\pm$ 0.01 & 2.84  $\pm$ 0.01\% &  74.46 $\pm$ 0.03\\
MAP-Elites (line)  & 3.75 $\pm$ 0.02 & 5.45  $\pm$ 0.03\%  &  76.25 $\pm$ 0.04\\
CMA-ME & 4.92 $\pm$ 0.04 & 7.26 $\pm$ 0.05\% & 76.15 $\pm$ 0.08\\
\bottomrule
\end{tabular}}
\vspace{1em}
\caption{Results for each derivative-free QD algorithm evaluated on the linear projection domains. These experiments were run with archives of resolution $500 \times 500$ instead of $100 \times 100$.} 
\label{tab:cmame_extra}
\end{table}

\section{Additional Results} 
\subsection{Generated Archives and Additional Metrics.} 

\begin{figure*}[t!]
\includegraphics[width=1.0\columnwidth]{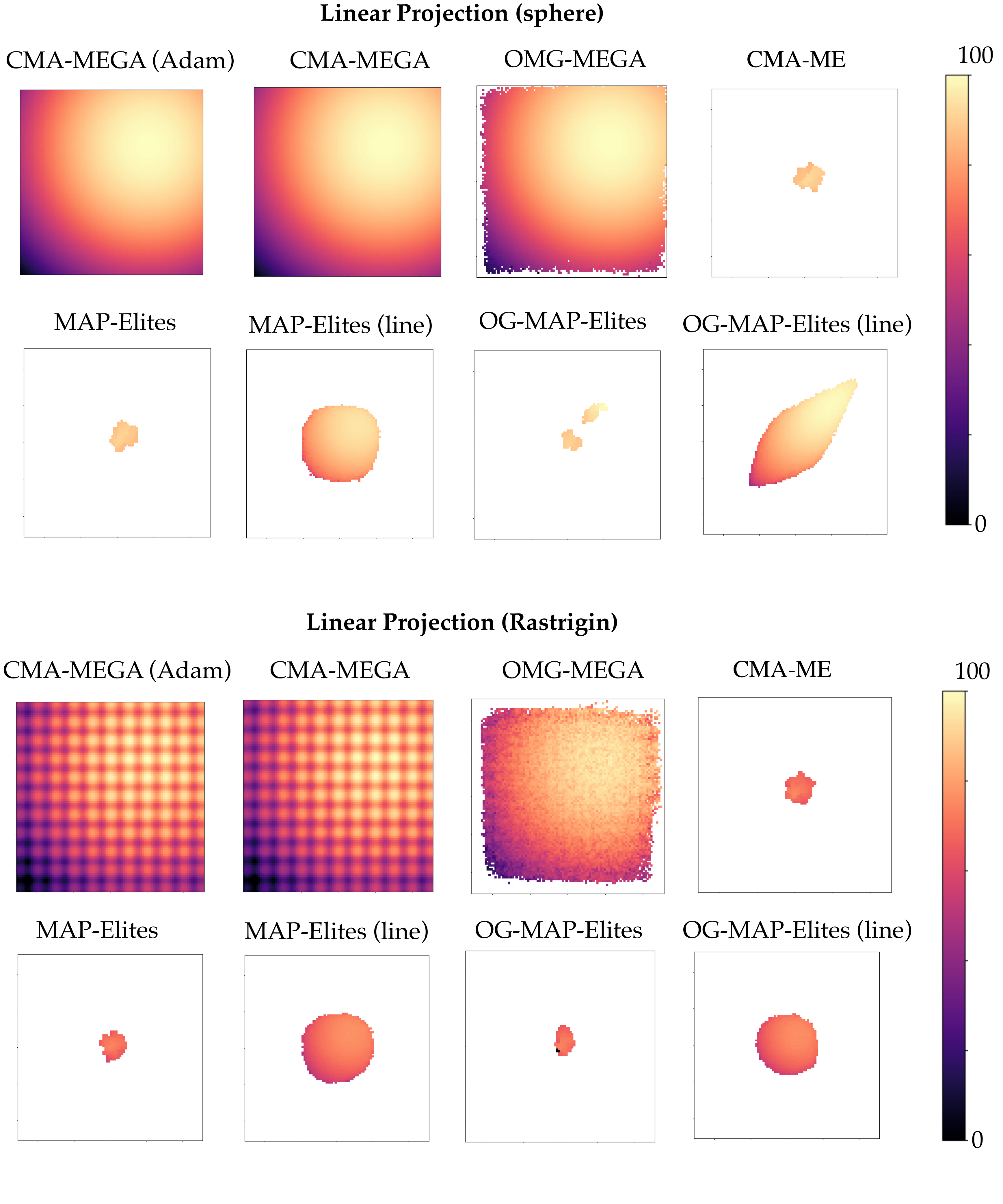}
\caption{Example archives for each algorithm evaluated in the linear projection domain.}
\label{fig:results}
\end{figure*}

\begin{figure*}[t!]
\includegraphics[width=1.0\columnwidth]{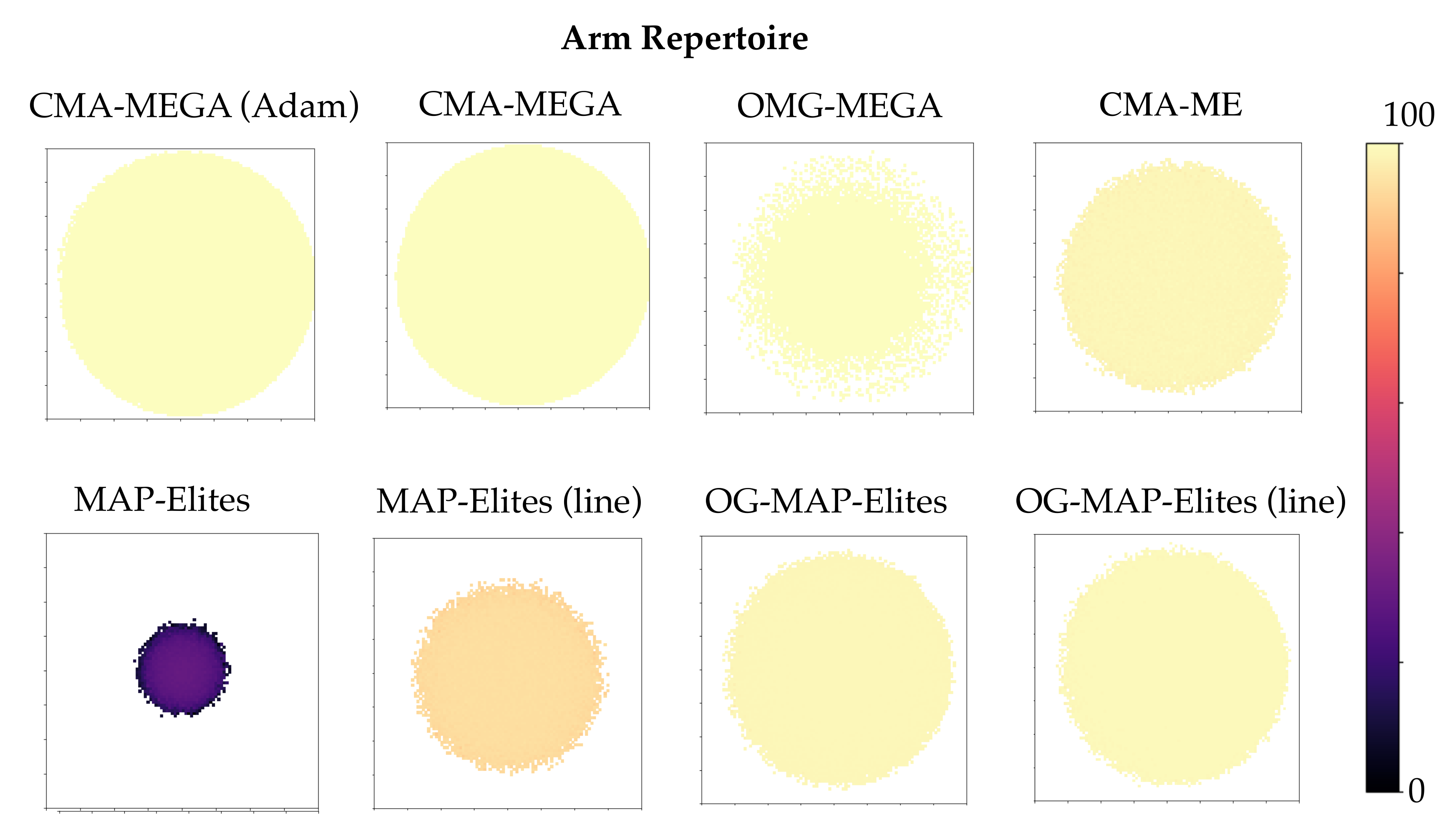}
\caption{Example archives for each algorithm evaluated in the arm repertoire domain.}
\label{fig:results}
\end{figure*}

\begin{figure*}[t!]
\includegraphics[width=1.0\columnwidth]{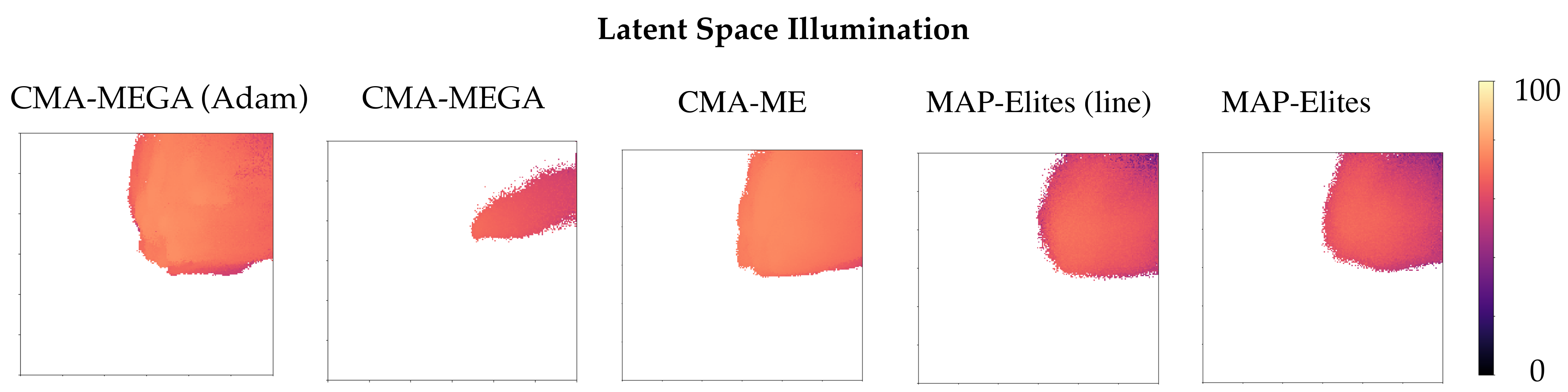}
\caption{Example archives for each algorithm in the latent space illumination (LSI) domain.}
\label{fig:results}
\end{figure*}

\begin{table*}[t]
\centering
\resizebox{0.7\linewidth}{!}{
\begin{tabular}{l|rrr}
\hline
             & \multicolumn{3}{c}{LP (sphere)}   \ \\ 
    \toprule
Algorithm     & QD-score & Coverage  & Best \\
    \midrule
MAP-Elites  &  1.04 $\pm$ 0.03& 1.17 $\pm$ 0.04\%  &  90.60 $\pm$ 0.03\\
MAP-Elites$^*$  &  5.19 $\pm$ 0.05 & 5.93 $\pm$ 0.06$\%$ &  91.39 $\pm$ 0.02 \\
MAP-Elites (line)  & 12.21 $\pm$ 0.06& 14.32 $\pm$ 0.07\%  &  94.89 $\pm$ 0.04\\
MAP-Elites (line)$^*$  & 34.74  $\pm$ 0.05 & 41.31 $\pm$ 0.06$\%$  &  98.83 $\pm$ 0.01\\
CMA-ME    & 1.08  $\pm$ 0.03 & 1.21 $\pm$ 0.04\% & 91.66 $\pm$ 0.01\\
CMA-ME$^*$ &  5.30   $\pm$ 0.04&  6.03 $\pm$ 0.05\%& 92.01 $\pm$ 0.02\\
 OG-MAP-Elites  & 1.52 $\pm$ 0.09& 1.67 $\pm$ 0.10\% & \textbf{100.00} $\pm$ \textbf{0.00}\\
 OG-MAP-Elites (line) & 15.01 $\pm$ 0.06 & 17.41 $\pm$ 0.08\% & \textbf{100.00} $\pm$ \textbf{0.00} \\
OMG-MEGA   & 71.58 $\pm$ 0.10 & 92.09 $\pm$ 0.21\% & \textbf{100.00} $\pm$ \textbf{0.00} \\
CMA-MEGA   & 75.29 $\pm$ 0.00 & \textbf{100.00} $\pm$ \textbf{0.00}\% & \textbf{100.00} $\pm$ \textbf{0.00}\\
CMA-MEGA (Adam)   & \textbf{75.30} $\pm$ \textbf{0.00} & \textbf{100.00} $\pm$ \textbf{0.00}\% & \textbf{100.00}  $\pm$ \textbf{0.00}\\
  \bottomrule
\end{tabular}
}
\resizebox{0.7\linewidth}{!}{
\begin{tabular}{l|rrr}
             & \multicolumn{3}{c}{LP (Rastrigin)}   \ \\ 
    \toprule
Algorithm     & QD-score & Coverage  & Best \\
    \midrule
MAP-Elites  &  1.18 $\pm$ 0.02 & 1.72 $\pm$ 0.04\%  &  74.48 $\pm$ 0.06\\
MAP-Elites$^*$  & 3.89 $\pm$ 0.03 & 5.96 $\pm$ 0.05$\%$ &  74.60 $\pm$ 0.04 \\
MAP-Elites (line)  & 8.12 $\pm$ 0.03 & 11.79 $\pm$ 0.05\%  &  77.43 $\pm$ 0.05\\
MAP-Elites (line)$^*$  &  22.65 $\pm$ 0.05 & 33.19 $\pm$ 0.08$\%$ &  81.13 $\pm$ 0.03\\
CMA-ME    & 1.21 $\pm$ 0.02 & 1.76 $\pm$ 0.03\% & 75.61 $\pm$ 0.05\\
CMA-ME$^*$ &  4.04 $\pm$ 0.02 &  6.13 $\pm$ 0.03\% &  75.71 $\pm$ 0.03\\
 OG-MAP-Elites  & 0.83 $\pm$ 0.03& 1.26 $\pm$ 0.03\% & 74.45 $\pm$ 0.03\\
 OG-MAP-Elites (line) & 6.10 $\pm$ 0.05& 8.85 $\pm$ 0.07\% & 76.61 $\pm$ 0.06  \\
OMG-MEGA   & 55.90 $\pm$ 0.21 & 77.00 $\pm$ 0.34\% & 97.55 $\pm$ 0.06 \\
CMA-MEGA   & 62.54 $\pm$ 0.00 & \textbf{100.00} $\pm$ \textbf{0.00}\% & 99.98 $\pm$ 0.00\\
CMA-MEGA (Adam)   & \textbf{62.58} $\pm$ \textbf{0.00} & \textbf{100.00} $\pm$ \textbf{0.00}\% & \textbf{99.99} $\pm$ \textbf{0.00}\\
  \bottomrule
\end{tabular}
}
\resizebox{0.7\linewidth}{!}{
\begin{tabular}{l|rrr}
             & \multicolumn{3}{c}{Arm Repertoire}   \ \\ 
    \toprule
Algorithm     & QD-score & Coverage  & Best \\
    \midrule
MAP-Elites  &  1.97 $\pm$ 0.05 & 8.06 $\pm$ 0.11\%    &  31.64 $\pm$ 0.44\\
MAP-Elites$^*$ & 19.43 $\pm$ 0.30  &  27.17 $\pm$ 0.35$\%$ &   75.42 $\pm$ 0.08 \\
MAP-Elites (line)  & 33.51 $\pm$ 0.65 & 35.79 $\pm$ 0.66\%  &  94.57 $\pm$ 0.10\\
MAP-Elites (line)$^*$ & 61.59  $\pm$ 0.04 & 62.73 $\pm$ 0.04$\%$ &  98.40 $\pm$ 0.00 \\
CMA-ME    & 55.98 $\pm$ 0.60 & 56.95 $\pm$ 0.61\% & 99.10 $\pm$ 0.00\\
CMA-ME$^*$  & 61.45 $\pm$ 0.22 & 62.48 $\pm$ 0.22\% &  99.10 $\pm$ 0.00 \\
 OG-MAP-Elites  & 57.17 $\pm$ 0.04 & 58.08 $\pm$ 0.04\% & 98.64 $\pm$ 0.00\\
 OG-MAP-Elites (line) & 59.66 $\pm$ 0.03 & 60.28 $\pm$ 0.03\%& 99.17 $\pm$ 0.00 \\
OMG-MEGA   & 44.12 $\pm$ 0.06 & 44.13 $\pm$ 0.06\% & \textbf{100.00} $\pm$ \textbf{0.00} \\
CMA-MEGA   & \textbf{74.18} $\pm$ \textbf{0.15} & \textbf{74.18} $\pm$ \textbf{0.15}\% & \textbf{100.00} $\pm$ \textbf{0.00}\\
CMA-MEGA (Adam)   & 73.82 $\pm$ 0.20 & 73.82 $\pm$ 0.20\% & \textbf{100.00} $\pm$ \textbf{0.00}\\
  \bottomrule
\end{tabular}
}

\resizebox{0.7\linewidth}{!}{
\begin{tabular}{l|rrr}
             & \multicolumn{3}{c}{LSI}   \ \\ 
    \toprule
Algorithm     & QD-score & Coverage  & Best \\
    \midrule
MAP-Elites  &  13.88 $\pm$ 0.11 & 23.15 $\pm$ 0.14\%    &  69.76 $\pm$ 0.07\\
MAP-Elites (line)  & 16.54 $\pm$ 0.28& 25.73 $\pm$ 0.31\%  &  72.63 $\pm$ 0.28\\
CMA-ME    & 18.96 $\pm$ 0.17 & 26.18 $\pm$ 0.24\% & 75.84 $\pm$ 0.10\\
CMA-MEGA   & 5.36 $\pm$ 0.78 & 8.61 $\pm$ 1.19\% & 68.74 $\pm$ 1.20\\
CMA-MEGA (Adam)   & \textbf{21.82} $\pm$ \textbf{0.18} & \textbf{30.73} $\pm$ \textbf{0.15}\% & \textbf{76.89} $\pm$ \textbf{0.15}\\
  \bottomrule
\end{tabular}
}
\caption{Results: Values with standard errors of QD-score, coverage, and best solution after 10,000 iterations for each algorithm and domain. Starred algorithms are run for $10^6$ iterations.
 }
\label{tab:results_detailed}
\end{table*}

 \begin{figure}[!t]
\centering
\includegraphics[width=1.0\linewidth]{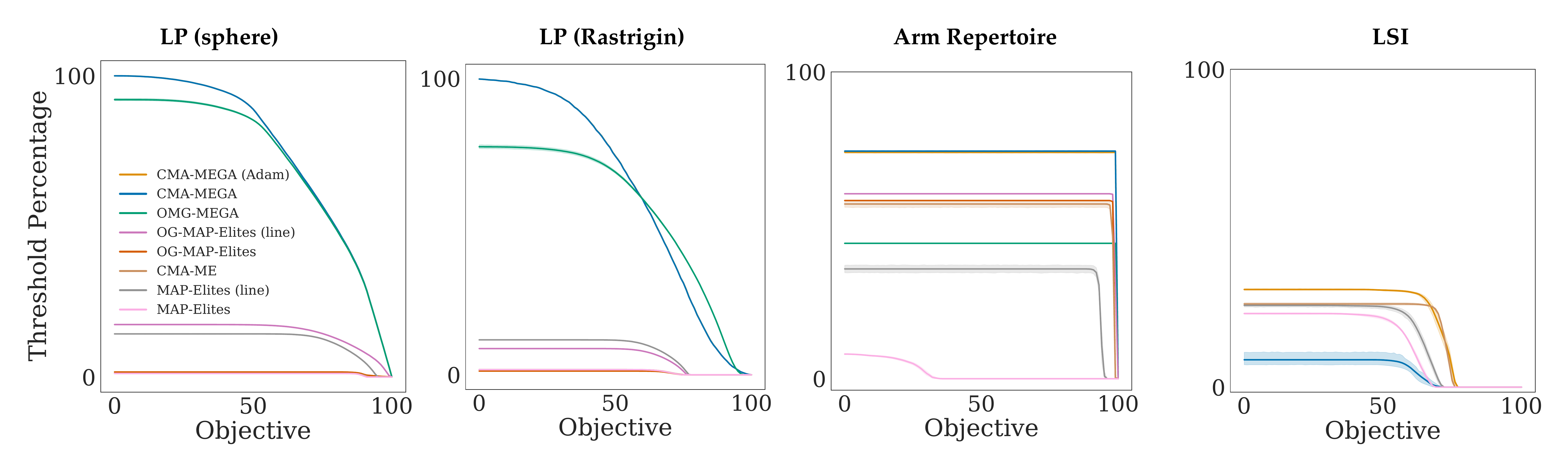}
\caption{Percentage of cells in the y-axis with objective values larger than or equal to a threshold specified in the x-axis. The percentage is the number of filled cells (filtered by the threshold) over the archive size. Larger area under each curve indicates better performance.}
\label{fig:cumulative_plots}
\end{figure}

 Fig.~\ref{fig:results} presents example archives for each algorithm and domain combination. Table~\ref{tab:results_detailed} presents the values of the QD-score, coverage, and best solution for each algorithm and domain.\footnote{We note that the QD-score and coverage values of OG-MAP-Elites are slightly different between Table~\ref{tab:results} and Table~\ref{tab:results_detailed}. Table~\ref{tab:results_detailed} contains the correct values and we will update Table~\ref{tab:results} in the revised version. The change does not affect the tests for statistical significance or any of our quantitative and qualitative findings of the experiments in section~\ref{sec:experiments}.} We additionally ran MAP-Elites, MAP-Elites (line), and CMA-ME for $10^6$ iterations, 20 trials, in the linear projection and arm repertoire domains, and present the results as \mbox{MAP-Elites$^*$}, \mbox{MAP-Elites (line)$^*$}, and \mbox{CMA-ME$^*$}. We observe that running CMA-MEGA for only 10,000 iterations outperforms each algorithm in both domains. 

We note that the QD-score metric combines both the number of cells filled in the archive (diversity) and the objective value of each occupant (quality). To disambiguate the two, we show in Fig.~\ref{fig:cumulative_plots} the percentage of cells (y-axis) that have objective value greater than the threshold specified in the x-axis.

 \begin{figure}[!t]
\centering
\includegraphics[width=0.9\linewidth]{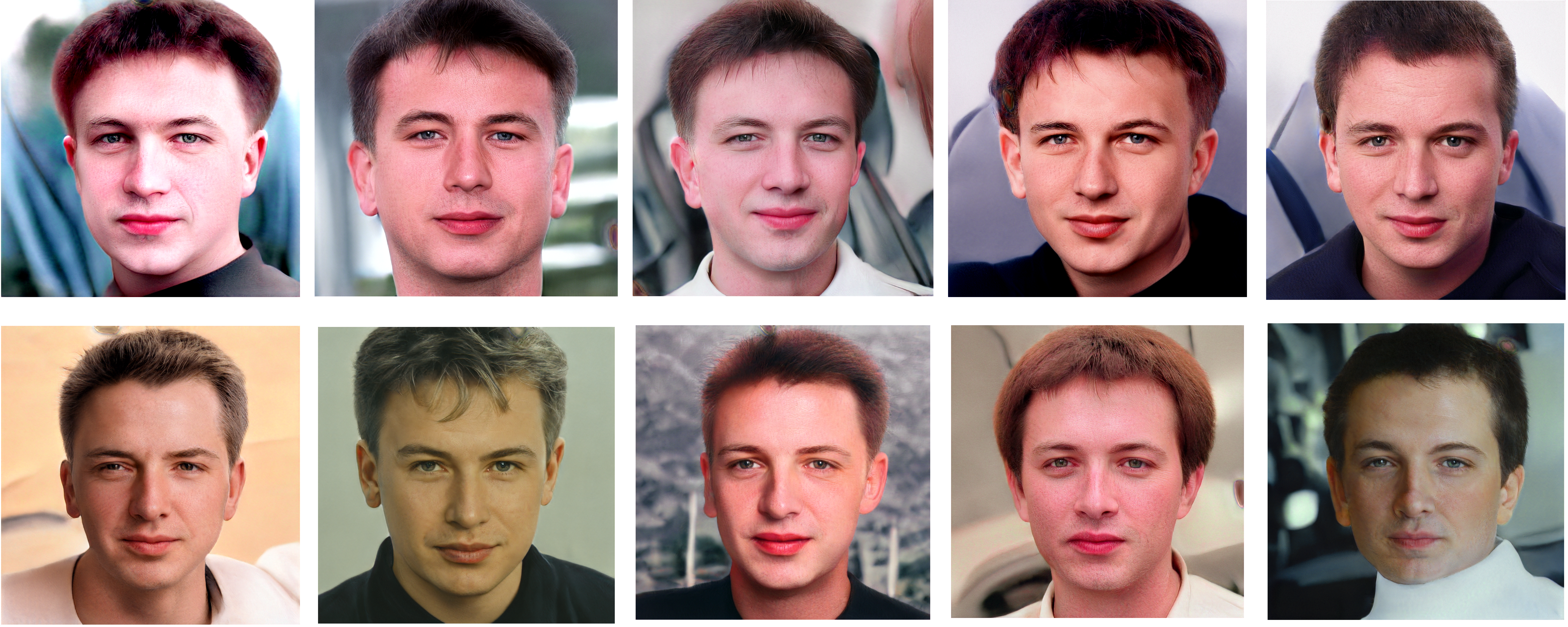}
\caption{Images generated with the single-objective Adam optimizer (top) and with CMA-MEGA (bottom) on the StyleGAN+CLIP model. Each algorithm optimizes the objective prompt ``Elon Musk with short hair.''}
\label{fig:images_comparison}
\end{figure}

\begin{figure}
\centering
\begin{tabular}{cc}
\begin{subfigure}{0.45\textwidth}
    \includegraphics[width=\linewidth]{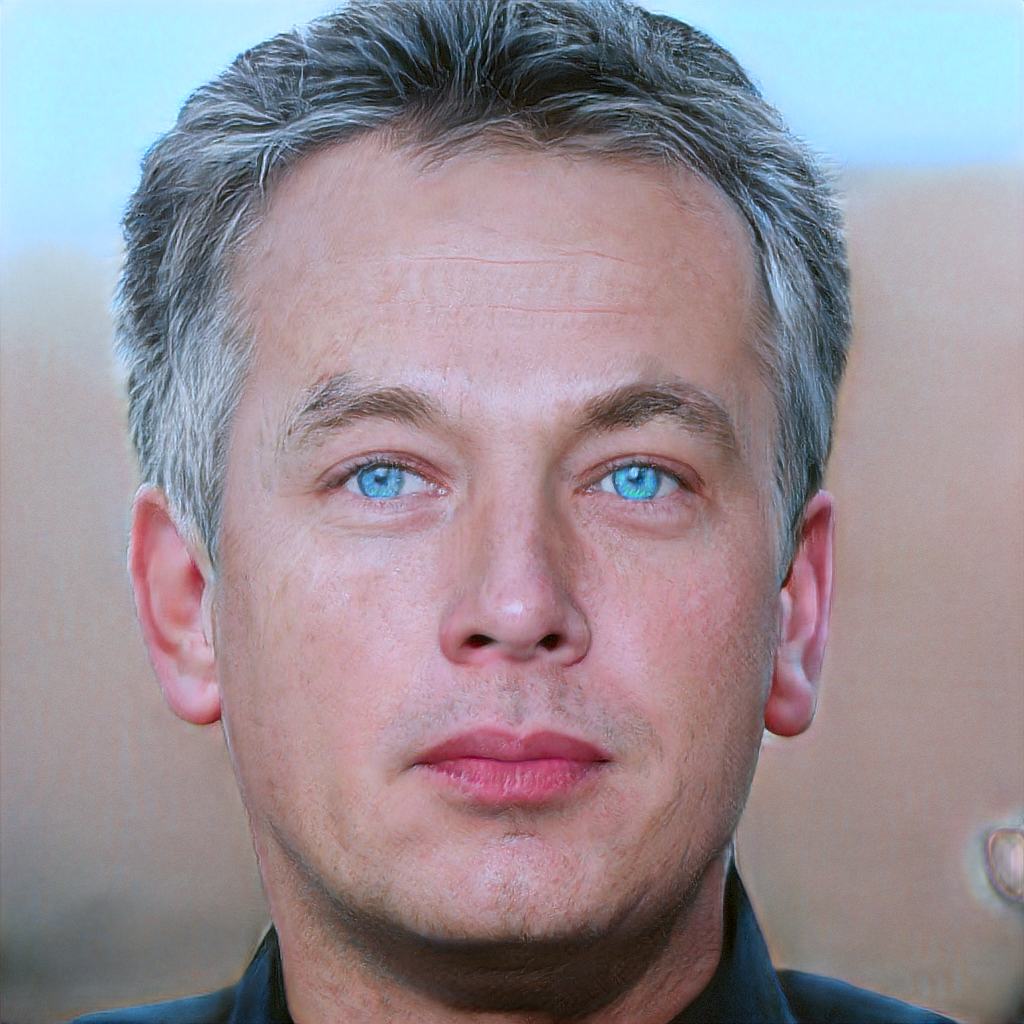}
    \caption{}
    \label{fig:my_label}
\end{subfigure}&
\begin{subfigure}{0.45\textwidth}
    \includegraphics[width=\linewidth]{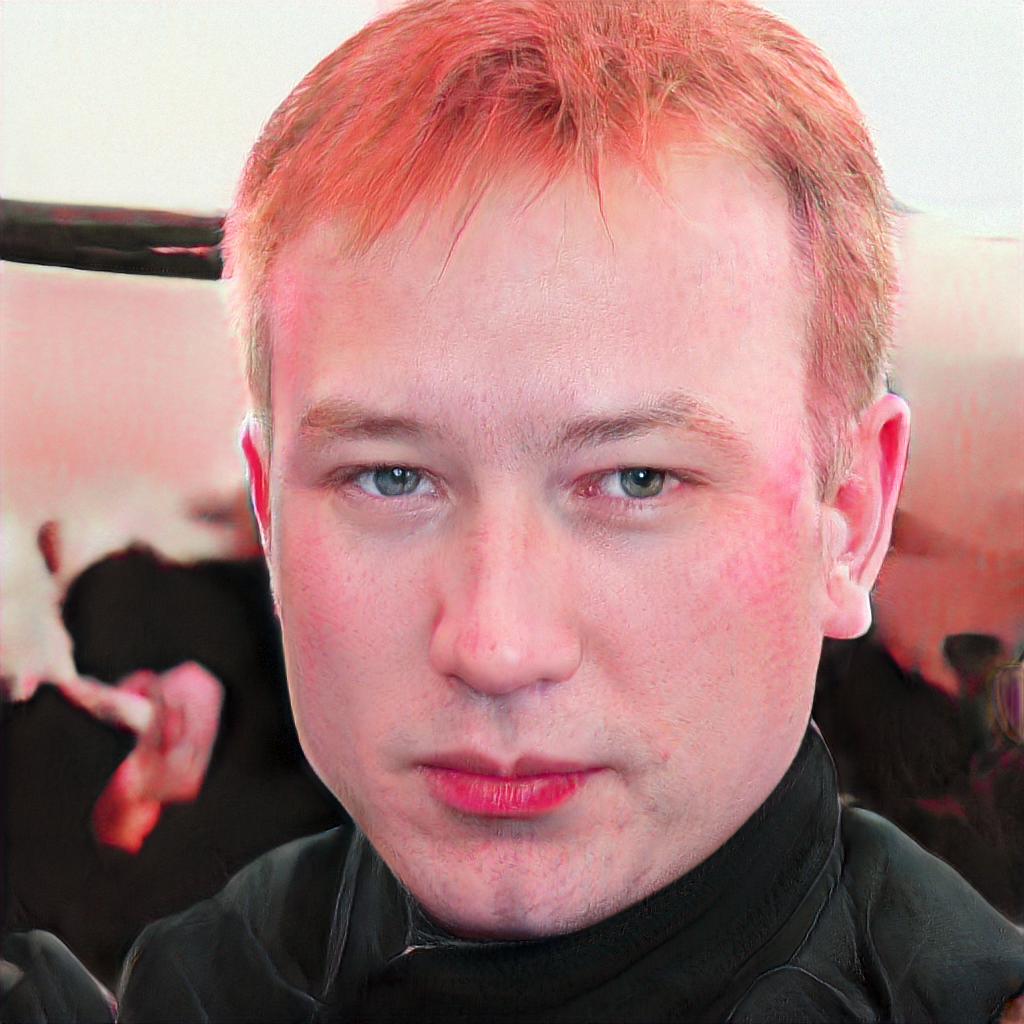}
    \caption{}
    \label{fig:my_label}
\end{subfigure}\\
\end{tabular}
\caption{Cherry-picked intermediate images generated from CMA-MEGA (Adam) during the LSI experiments for the prompt ``Elon Musk with short hair.''. These images were discovered intermediate solutions, but did not end up in the final archive returned by CMA-MEGA (Adam).}
\label{fig:cherry}
\end{figure}

\noindent\textbf{Qualitative Results.} 
We can also compare the quality of the generated images with latent space illumination, with the images generated when using a single-objective Adam optimizer, where we optimize StyleGAN+CLIP~\cite{styleclip} with Adam as a baseline (instead of running a QD algorithm). We run 5 different trials for 10,000 iterations, and for each trial we present the image that the algorithm converged to. StyleGAN+CLIP with Adam gradually drifted towards unrealistic images, so we excluded from our selection images that were unrealistic or with large artifacts. We used the same prompt as the objective of section~\ref{sec:experiments}: ``Elon Musk with short hair.''. We compare these images with the best image, according to the objective, of each of the 5 generated archives of CMA-MEGA (Adam). We used in both conditions the Adam implementation of ESTool (see accompanying source code) with the same hyperparameters. We observe that the quality of the images is comparable (Fig.~\ref{fig:images_comparison}). 

While running the LSI experiments, we discovered that several intermediate solutions were of better visual quality than the final images present in the generated collage. Fig.~\ref{fig:cherry} shows two such images cherry-picked from all intermediate solutions generated by CMA-MEGA (Adam) during the experiments. According to CLIP, these images are of lower quality than the images present in the collage of Fig.~\ref{fig:collage} in the main paper. We interpret this finding as CMA-MEGA overfitting to the CLIP loss function. Several methods exist in the generative art community for mitigating this affect. For example, CLIPDraw~\citep{frans2021clipdraw} augments the generated image via randomized data augmentations then evaluates the generated batch with CLIP. The average loss of the entire batch replaces evaluating a single generated image with CLIP. A similar method could be applied to our LSI approach to improve the perceptual quality of generated images. 

\subsection{Additional Experiments in the LSI Domain}

We include additional experiments in the LSI domain to assess DQD's qualitative and quantitative performance on different prompts. The two additional experiments were run with the objective prompts ``A photo of Beyonce'' and a ``A photo of Jennifer Lopez'' with the measure prompts ``A small child.'' and ``A woman with long blonde hair.''.

Table~\ref{tab:lsi_additional_runs} presents quantitative metrics of the additional runs as well as the Elon Musk experiment from the main paper. We observe that CMA-MEGA (Adam) retains top quantitative performance against each baseline for each experiment.

\begin{figure}[!t]
\centering
\includegraphics[width=1.0\linewidth]{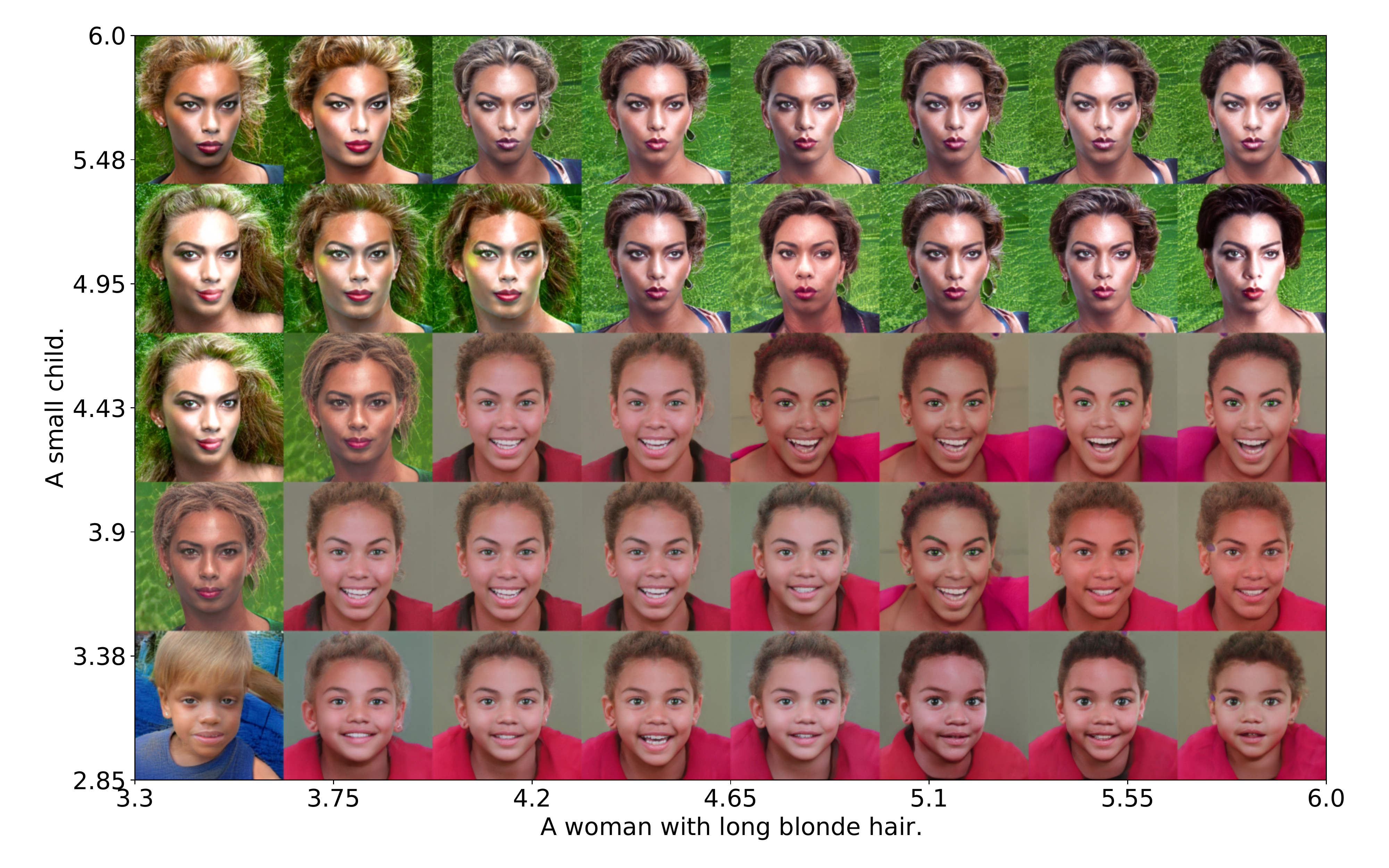}
\caption{Result of latent space illumination for objective ``A photo of Beyonce.'' and for measures ``A small child.'' and ``A woman with long blonde hair.''. The axes values indicate the score returned by the CLIP model, where lower score indicates a better match.}

\label{fig:collage_Beyonce}
\end{figure}

\begin{figure}[!t]
\centering
\includegraphics[width=1.0\linewidth]{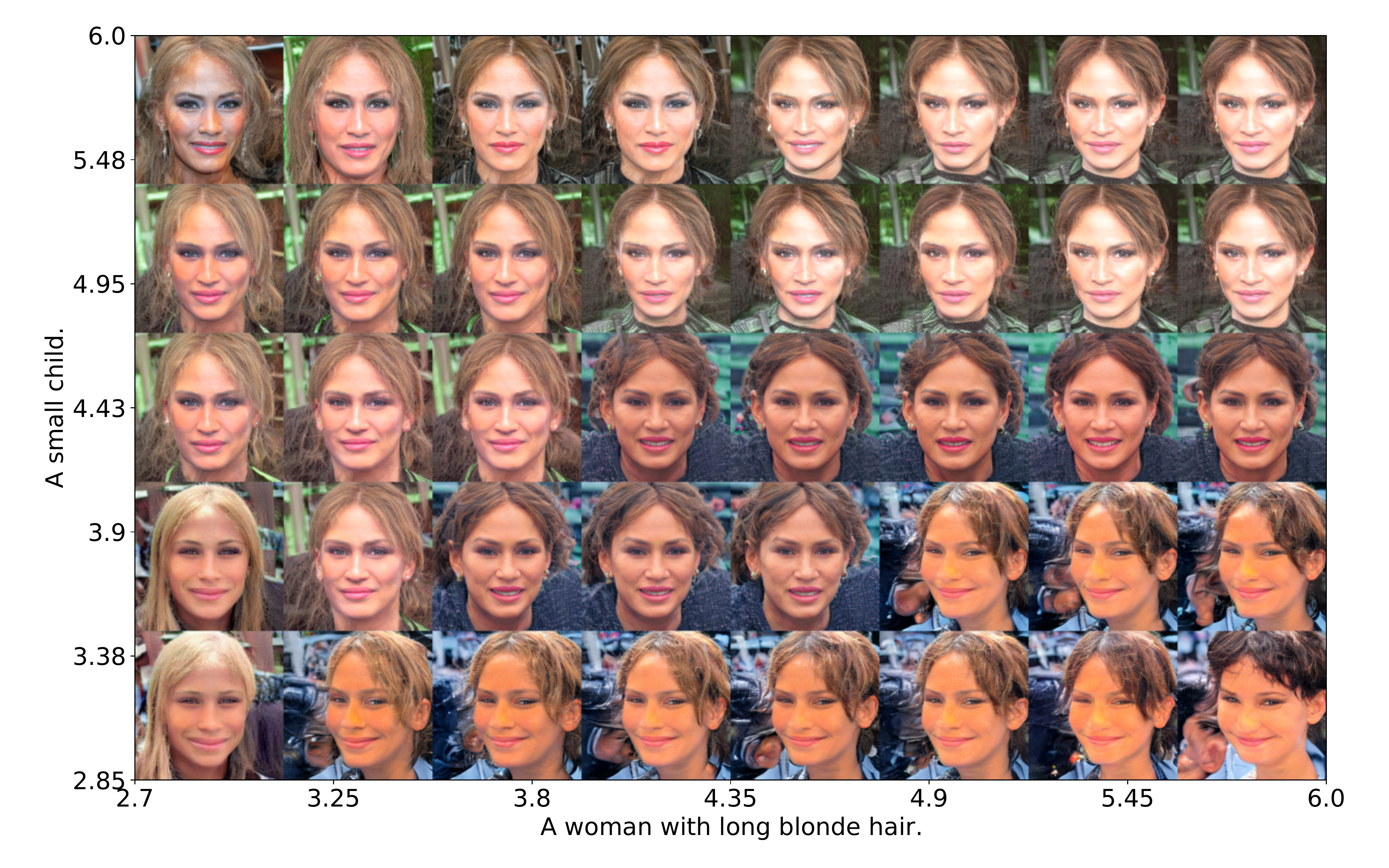}
\caption{Result of latent space illumination for objective ``A photo of Jennifer Lopez.'' and for measures ``A small child.'' and ``A woman with long blonde hair.''. The axes values indicate the score returned by the CLIP model, where lower score indicates a better match.}

\label{fig:collage_Lopez}
\end{figure}

Fig.~\ref{fig:collage_Beyonce} and Fig.~\ref{fig:collage_Lopez} contain collages for the Beyonc\'e and Jennifer Lopez experiments, respectively. For the Jennifer Lopez and Elon Musk experiments, we report that all archive collages were of similar quality. However, for Beyonc\'e only one of the collages, the one presented in Fig.~\ref{fig:collage_Beyonce}, did not have significant artifacts. CLIP guided CMA-MEGA (Adam) far out of the latent Gaussian distribution for the remaining runs of the Beyonc\'e prompt and became confident in low quality images that overwrote previously discovered high quality images in the archive. We highlight the concern that CLIP may perform differently for text prompts identifying celebrities of different ethnic backgrounds. 

\begin{figure}[!t]
\centering
\includegraphics[width=1.0\linewidth]{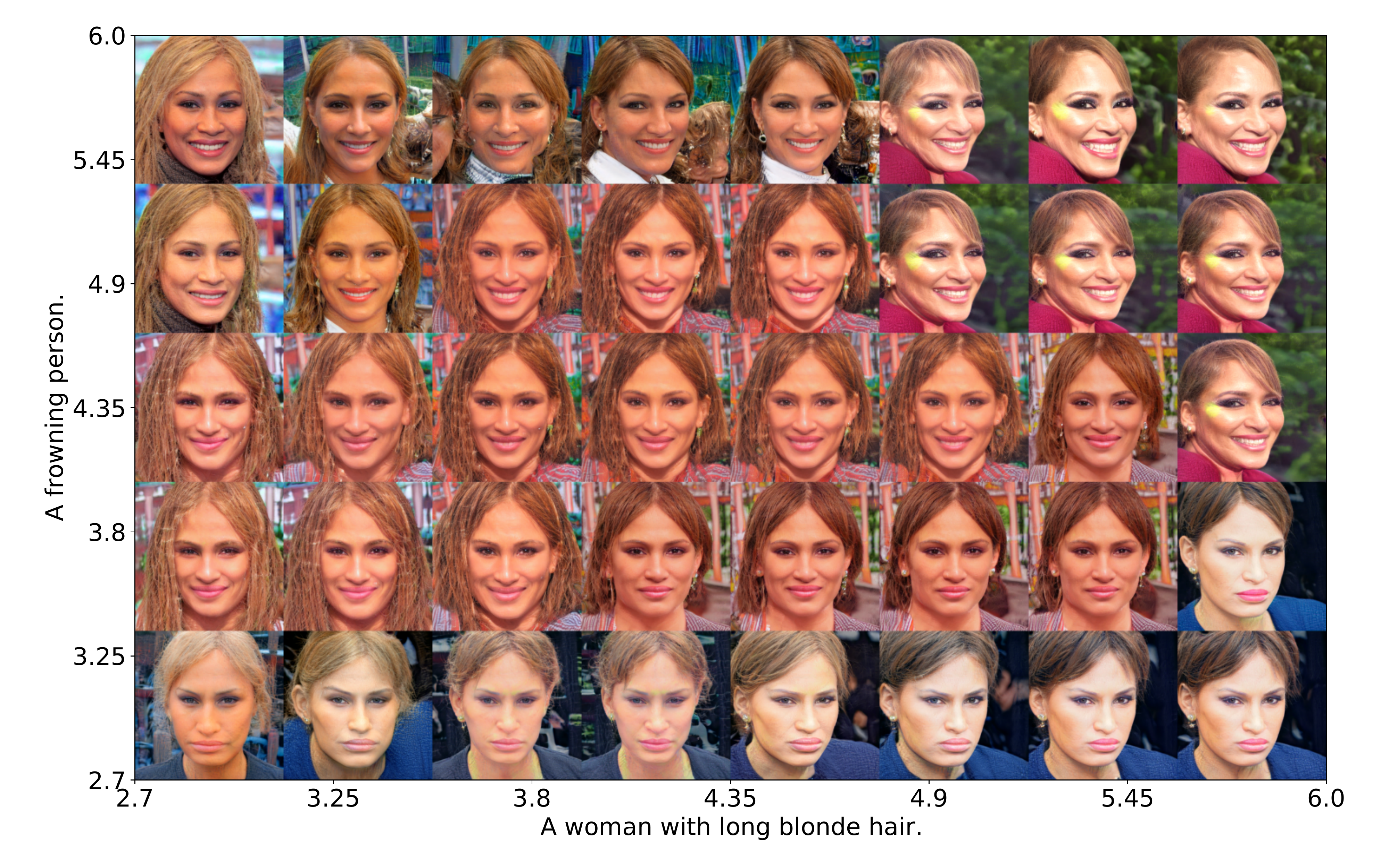}
\caption{Result of latent space illumination for objective ``A photo of Jennifer Lopez.'' and for measures ``A frowning person.'' and ``A woman with long blonde hair.''. The axes values indicate the score returned by the CLIP model, where lower score indicates a better match.}

\label{fig:collage_Lopez2}
\end{figure}

For Beyonc\'e, the collage contains faces of different ages for the measure ``A small child'', a proxy prompt for ``age''. However, for the Jennifer Lopez prompt, DQD constructs younger faces of Jennifer Lopez by places an older looking face on a younger looking head. We include an additional qualitative run of CMA-MEGA (Adam) on Jennifer Lopez (see Fig.~\ref{fig:collage_Lopez2}), where we replace the measure prompt ``A small child.'' with the prompt ``A frowning person''. In this prompt, we find faces of a younger Jennifer Lopez not discovered in Fig.~\ref{fig:collage_Lopez}. This result shows that younger faces of Jennifer Lopez can be generated by StyleGAN and identified by CLIP, but not discovered by DQD when the measure function approximated ``age''. We hypothesize that this is a limitation of using a proxy measure for age, rather than a predictive model that can directly measure the abstract concept of age.

\begin{table}[t!]
\centering
\resizebox{0.7\linewidth}{!}{
\begin{tabular}{l|rrr}
             & \multicolumn{3}{c}{LSI: Elon Musk}   \ \\ 
    \toprule
Algorithm     & QD-score & Coverage  & Best \\
    \midrule
MAP-Elites  &  13.88 $\pm$ 0.11 & 23.15 $\pm$ 0.14\%    &  69.76 $\pm$ 0.07\\
MAP-Elites (line)  & 16.54 $\pm$ 0.28& 25.73 $\pm$ 0.31\%  &  72.63 $\pm$ 0.28\\
CMA-ME    & 18.96 $\pm$ 0.17 & 26.18 $\pm$ 0.24\% & 75.84 $\pm$ 0.10\\
CMA-MEGA   & 5.36 $\pm$ 0.78 & 8.61 $\pm$ 1.19\% & 68.74 $\pm$ 1.20\\
CMA-MEGA (Adam)   & \textbf{21.82} $\pm$ \textbf{0.18} & \textbf{30.73} $\pm$ \textbf{0.15}\% & \textbf{76.89} $\pm$ \textbf{0.15}\\
  \bottomrule
\end{tabular}
}

\resizebox{0.7\linewidth}{!}{
\begin{tabular}{l|rrr}
             & \multicolumn{3}{c}{LSI: Beyonce}   \ \\ 
    \toprule
Algorithm     & QD-score & Coverage  & Best \\
    \midrule
MAP-Elites  &  12.84 $\pm$ 0.10 & 19.41 $\pm$ 0.16\%  &   71.42 $\pm$ 0.14\\
MAP-Elites (line)  &  14.40 $\pm$ 0.09&  21.11 $\pm$ 0.16\%  &  73.04 $\pm$ 0.05\\
CMA-ME & 14.00 $\pm$ 0.62  & 19.57 $\pm$ 0.90\% &  74.11 $\pm$ 0.08 \\
CMA-MEGA  &  3.37 $\pm$ 0.44 & 6.38 $\pm$ 0.71\% & 59.96 $\pm$ 1.40 \\
CMA-MEGA (Adam) & \textbf{16.08} $\pm$ \textbf{0.37} & \textbf{22.58} $\pm$ \textbf{0.57}\% &  \textbf{74.95} $\pm$ \textbf{0.27} \\
  \bottomrule
\end{tabular}
}

\resizebox{0.7\linewidth}{!}{
\begin{tabular}{l|rrr}
             & \multicolumn{3}{c}{LSI: Jennifer Lopez}   \ \\ 
    \toprule
Algorithm     & QD-score & Coverage  & Best \\
    \midrule
MAP-Elites  &  12.51 $\pm$ 0.28 & 19.18 $\pm$ 0.48 \%    &  70.87 $\pm$ 0.27 \\
MAP-Elites (line)  &  14.73 $\pm$ 0.06 & 21.60 $\pm$ 0.08\%  &  73.50 $\pm$ 0.13\\
CMA-ME    &  15.24 $\pm$ 0.37 & 20.86 $\pm$ 0.50\% & 75.39 $\pm$ 0.09\\
CMA-MEGA   &  3.14 $\pm$ 0.14 & 5.58 $\pm$ 0.13\% &  64.88 $\pm$ 2.89 \\
CMA-MEGA (Adam)   &  \textbf{17.06} $\pm$  \textbf{0.10} & \textbf{23.40} $\pm$ \textbf{0.14}\% &  \textbf{76.02} $\pm$ \textbf{0.08}\\
  \bottomrule
\end{tabular}
}~\\
\caption{Results from additional runs for Jennifer Lopez and Beyonc\'e. For comparison purposes, we include the original results for Elon Musk in the top table.} 
\label{tab:lsi_additional_runs}
\end{table}



\end{document}